\documentclass[manuscript,screen,review]{article}
\usepackage{xcolor}
\usepackage{arxiv}
\usepackage[utf8]{inputenc}
\usepackage{graphicx}
\usepackage{float}
\usepackage{subfigure}
\usepackage{amsmath}

\usepackage[utf8]{inputenc} 
\usepackage[T1]{fontenc}    
\usepackage{hyperref}       
\usepackage{url}            
\usepackage{booktabs}       
\usepackage{amsfonts}       
\usepackage{nicefrac}       
\usepackage{microtype}      
\usepackage{lipsum}
\usepackage{graphicx}

\usepackage[algoruled, linesnumbered]{algorithm2e}
\usepackage{xcolor}
\usepackage{enumitem}
\usepackage{fixmath}

\AtBeginDocument{%
  \providecommand\BibTeX{{%
    \normalfont B\kern-0.5em{\scshape i\kern-0.25em b}\kern-0.8em\TeX}}}
    
\newcommand{\ii}{\textit}
\newcommand{\bb}{\textbf}

\newcommand{\mb}{\mathbold}
\usepackage{ulem}

\title{Forgetful Forests: high performance learning data structures  for streaming data under concept drift}

\author{
 Zhehu Yuan \\
  Courant Institute of Mathematical Sciences\\
  New York University\\
  New York, N.Y. 10012 \\
  \texttt{zy2262@nyu.edu} \\
   \And
 Yinqi Sun\\
  Courant Institute of Mathematical Sciences\\
  New York University\\
  New York, N.Y. 10012 \\
  \texttt{ys3540@nyu.edu} \\
  \And
 Dennis Shasha \\
  Courant Institute of Mathematical Sciences\\
  New York University\\
  New York, N.Y. 10012 \\
  \texttt{shasha@cims.nyu.edu} \\
}

\begin{document}
\maketitle
\begin{abstract}
Database research can help machine learning performance in many ways. One way is to design better data structures. This paper combines the use of incremental computation and sequential and probabilistic filtering to enable "forgetful" tree-based learning algorithms to cope with concept drift data (i.e., data whose function from input to classification changes over time).

The forgetful algorithms described in this paper achieve high time performance while maintaining high quality predictions on streaming data. Specifically, the algorithms are up to $24$ times faster than state-of-the-art incremental algorithms with at most a $2$\% loss of accuracy, or at least $2$ times faster without any loss of accuracy. This makes such structures suitable for high volume streaming applications.
\end{abstract}

{\bf Keywords}: Concept Drift, Machine Learning, Incremental Algorithms, Tree Data Structures

\maketitle
\section{Introduction}

\ \\Supervised machine learning\cite{Cunningham2008} tasks start with a set of labeled data. Researchers partition that data into training data and test data. They train their favorite algorithms on the training data and then derive accuracy results on the test data. The hope is that these results will hold on to yet-to-be-seen data because the mapping between input data and output label (for classification tasks) doesn't change.

\ \\This paradigm works well for applications such as medical research. In such settings, if a given set of lab results $L$ indicate a certain diagnosis $d$ at time $t$, then that same set of input measurements $L$  will suggest diagnosis $d$ at a new time $t'$. 

\ \\However, there are many applications where the function between input and output label changes:   inflation rates, variants of epidemics, and  market forecasting. In such applications, the mapping from input to labeled output changes over time.  This requires more work, but also presents a time performance opportunity because a learning system can judiciously "forget" (i.e. discard) old data and  learn a new input-output function on only the relevant data and do so quickly.  
In addition to discarding data cleverly, such a system can take advantage of the properties of the data structures to speed up their maintenance.

These intuitions form the basic strategy of the forgetful data structures we describe here.

\section{Background}

\ \\The training process of many machine learning models take a set of training samples of the form $(x_1 , y_1), (x_2, y_2), ..., (x_N , y_N)$  where in each training sample $(\mb{x}_{i}, y_{i})\in (\mb{X}_{train}, \mb{Y}_{train})$, $\mb{x}_{i}$ is a vector of feature-values and ${y_{i}}$ is a class label\cite{DeepLearning, russel2010}. The goal is to learn a function from the $X$ values to the $y$ values. In the case when the mapping between $X$ and $y$ can change, an incremental algorithm will update the mapping as data arrives.  Specifically, after receiving the k-th batch of training data,   the \bb{parameters} of the model $f$ \bb{changes} to reflect that batch. At the end of the n-th training batch, the model $f_n$ can give a prediction of the following data point such that $\hat{y_{n+1}} = f_n (x_{n+1})$. This method of continuously updating the model on top of the previous model is called incremental learning.\cite{IncrementalNN} \cite{IncrementalSVM}
%

\ \\Conventional decision tree methods, like CART \cite{CART}, are not incremental. Instead, they learn a tree from an initial set of training data. Under the assumption that data is independent and identically distributed (i.i.d.), conventional decision tree methods form the tree once and for all. A naive incremental approach (needed when the data is not i.i.d.) would be to rebuild the tree from scratch periodically. But rebuilding the decision tree can be expensive. Alternative methods such as VFDT\cite{VFDT} or iSOUP-Tree \cite{Osojnik2017TreebasedMF} incrementally update the decision tree with the primary goal of reducing memory consumption.

%
\subsection{Hoeffding Tree}
In the Hoeffding Tree (or VFDT)\cite{10.1145/347090.347107}, each node  considers only a fixed subset of the training set, designated by a parameter $n$, and uses that data to choose the splitting attribute and value of that node. In this way, once  a node has been fitted on $n$ data points, it won't be updated anymore. The number of data points $n$ considered by each node is calculated using the Hoeffding bound \cite{doi:10.1080/01621459.1963.10500830}, $n=\frac{R^2 ln(1/\delta)}{2\epsilon^2}$, where $R$ is the range of the variable, $\delta$ is the confidence fraction which is set by user, and $\epsilon = \hat{G}(x_1)-\hat{G}(x_2)$ is the distance between the best splitter $x_1$ and the second best splitter $x_2$ based on the $\hat{G}$ function. For example, after a node has received $n$ training data points, we might have $\hat{G}(x_1)=0.2$ and $\hat{G}(x_2)=0.1$, where $\hat{G}(\dot)$ (e.g.,  information gain) is the heuristic measure used to choose spitting attributes,  and $x_1$ and $x_2$ are the best and second best splitting attributes for the $n$ training data points. Then for some fraction $\delta$, there will be a $1-\delta$ chance that $G(x_1)-G(x_2)>0.1-\epsilon$. 

%
\subsection{Adaptive Hoeffding Tree}

The Adaptive Hoeffding Tree \cite{10.1007/978-3-642-03915-7_22} 
will hold a variable-length window $W$ of recently seen data. We will have $1-\delta$ confidence that the splitting attribute has changed if the two windows are "large enough", and their heuristic measurement are "distinct enough". To define "large enough" and "distinct enough", the Adaptive Hoeffing tree uses  the  Hoeffding bound:, when $|\hat{G}(W)-\hat{G}(x_1)|$ is larger than $2*\epsilon$. In scikit-multiflow's implementation, a bootstrap sampling strategy is  applied to improve the algorithm's time performance.

%
\subsection{iSOUP-Tree}
In contrast to the Hoeffding Tree, the iSOUP-Tree \cite{Osojnik2017TreebasedMF} uses the FIMT-DD method \cite{Ikonomovska2011}. That works as follows. There are two learners at each leaf to make predictions. One learner is a linear function $y=wx+b$ used to predict the result, where $w$ and $b$ are variables trained with the data and the results that have already arrived at this leaf, $y$ is the prediction result, and $X$ is the input data. The  other learner computes the average value of the $y$ from the training data seen so far. The learner with the lower absolute error will be used to make predictions.  Different leaves in the same tree may choose different learners. 

\subsection{Adaptive Random Forest}
%
The Adaptive Random Forest \cite{AdaptiveRF} is a random forest, each of whose trees  is a Hoeffding Tree without early pruning. Whenever a new node is created, a random subset of features with a certain size is selected. Split attempts are limited to these features for the given node. The number of features in each subset is a hyperparameter that was set before training and will not change any more. The data is randomly selected with replacement from the training set, while the chance for each data to be selected is determined by a binomial distribution. The chance for a certain size of data to be selected is based on a Poisson distribution with $\lambda = 6$. To detect concept drift, it uses the the method of the Adaptive Hoeffding Tree as described above. Further, each tree has two confidence interval levels to reflect its poor performance in the face of concept drift.  When the lower confidence level of a tree is reached (meaning the tree has not been performing well), the Random Forest will create a new background tree that trains like any other  tree in the forest, but the background tree will not be contribute to the prediction. If the tree already has one background tree, the older one will be replaced. When the higher confidence level of a tree is reached (meaning the tree has been performing very badly), the Random Forest will delete that tree and replace it with the corresponding background tree. Adaptive Random Forests achieve state-of-the-art accuracy.

\section{Forgetful Data Structures}
\label{ForgetfulDataStructure}

This paper introduces both a Forgetful Decision Tree and a Forgetful Random Forest. They sequentially and probabilistically forget old data and combine the retained old data with new data to track datasets that may undergo concept drift. In the process, the values of several hyperparameters are adjusted depending on the relative accuracy of the data structure at hand (whether a decision tree or a random forest). We present the overall algorithms in subsections \ref{MainDT} and \ref{MainRF} but first  we present the subroutines, each of which takes several input parameters.

The {\bf accuracy} in this paper is measured based on the confusion matrix: the proportion of true positive plus true negative instances relative to all test samples:  
$$\frac{|True Positive ~ in ~ testset| + |True Negative ~ in ~testset|}{Size(testset)}$$ 

\subsection{Forgetful  decision trees}\label{forgetdecisiontrees}

\begin{algorithm}[H]
    \SetAlgoLined
	\SetKwInOut{Input}{input}
	\SetKwInOut{Global}{global}
	\SetKwInOut{Output}{output}
	\caption{BuildSubTree}\label{DT:BuildSubTree}
	
	\Input{ This is called when a subtree must be rebuilt from scratch.\\$\mb{X}$, $\mb{Y}$, incoming training data batch for current node
	\\ $E(\cdot)$, the stopping criteria, returns boolean 
        \\ $G(\cdot)$, a function to score the fitness of a feature for splitting.}
        \Global{
        $current.X, current.Y$, the data retained 
        \\ $current.gRecord$, the sorted data retained  
        \\ $current.ranges$, the ranges for splitting retained  
        \\ $current.label$, the prediction label  
        }
	\Begin{
        $current.X, current.Y \leftarrow \mb{X}, \mb{Y}$\;
        \uIf{E($current$, $current.X$, $current.Y$) is $True$}{
            $current.label \leftarrow$ the class having the greatest probability\;
        }
        \Else{
            $current.gRecord, ranges \leftarrow$ minG(X, Y)\;// find the best value ranges\\
            $Xs, Ys \leftarrow$ Split ($current.X$, $current.Y$) at $rangess$\;
            i=0\;
            \For{each $x, y$ in $Xs, Ys$}{
                $child \leftarrow BuildSubTree(x, y, current.children[i], E(.), G(.))$\;
                $i++$\;
            }
            $current.ranges \leftarrow ranges$\;
        }
    }
\end{algorithm}
\ \\
\begin{algorithm}[H]
        \SetAlgoLined
	\SetKwInOut{Input}{input}
	\SetKwInOut{Global}{global}
	\SetKwInOut{Output}{output}
	\caption{UpdateSubTree}
        \label{DT:UpdateSubTree}
	\Input{ $\mb{X}$, $\mb{Y}$, incoming batch of training data for current node
	\\ $E(\cdot)$, the stopping criteria 
	\\$G(\cdot)$, the function to score the fitness of a feature for splitting}
	\Global{
        $current.X, current.Y$, the data retained of previous update
        \\ $current.gRecord$, the sorted data retained of previous update
        \\ $current.ranges$, the ranges for splitting retained of previous update
        \\ $current.label$, the prediction label 
        }   
        \Begin{
        Discard the oldest $(size(X)+size(current.X)-currentParams.rSize)$ rows in $current.X$ and $current.Y$\;
        $current.X, current.Y \leftarrow current.X+X, current.Y+Y$\;// insert the incoming batch, $currentParams.rSize$ rows will be in $current$ after insertion.\\
        \label{algo:combineData}
        \uIf{E($current$, $current.X$, $current.Y$) is $True$} {
            // don't split subtree more\\
            $current.label \leftarrow$ the class having the greatest probability\;
        }
        \Else{
            $current.gRecord, ranges \leftarrow minGInc(X, Y, G, current.gRecord)$\;// find the best value ranges\\
            \For{each $range$ in $ranges$}{
                $child \leftarrow current.children[i]$\;
                \uIf{$range$ not in $current.ranges$}{
                    // we need to rebuild subtree\\
                    $x, y\leftarrow$ Split ($current.X$, $current.Y$) at $range$\; 
                    $child \leftarrow BuildSubTree(x,y,child, E(.), G(.))$\;
                }\Else{
                    $x, y\leftarrow$ Split ($X$, $Y$) at $range$\;
                    $child \leftarrow UpdateSubTree(x, y, child, E(.), G(.), child.(X, Y, gRecord, ranges, label)$, $currentParams.rSize)$\;
                }
                $i++$\;
            }
            $current.ranges \leftarrow ranges$\;
        }
    }
\end{algorithm}

\ \\ When new data is acquired from the data stream, function $UpdateSubTree(.)$ (Algorithm \ref{DT:UpdateSubTree}) will be called on the root node to incrementally and recursively update the entire decision tree.

\begin{itemize}
\item
The stopping criteria $E(\cdot)$ may comprise maximum tree height, minimum samples to split, or minimum impurity decrease. If it returns $True$, then the node is not further split. Also, if all data in some node have the same label, that node will not be further split.
\item
The evaluation function $G(\cdot)$ evaluates the score of each feature and each range for splitting. It will typically be a Gini Impurity\cite{gini} score or an entropy reduction coefficient. As we discuss below, the functions $minG$ and $minGInc$ find split points that minimize the weighted sum of scores of each subset of the data after splitting. Thus, the score must be evaluated on many split points (e.g., if the input attribute is $age$, then possible splitting criteria could be $age > 30$, $age > 32$, ...) to find the optimal ranges. 
Those evaluations run much faster if the data is sorted for each input attribute.
\item
$currentParams.rSize$ governs the size of $current.X+X$ and $current.Y+Y$ to be retained, when new data for $X$ and $Y$ arrives. For example, suppose $currentParams.rSize = 100$. Then $size(X)+size(current.X)-100$ of the oldest of $current.X$ and $current.Y$ (the data present before the incoming batch) will be discarded. The algorithm then appends $X$ and $Y$ to what remains of $current.X$ and $current.Y$.   All nodes in the subtrees will discard the same data items as the root node. In this way, the tree is trained with only the newest $currentParams.rSize$ of data in the tree. Discarding old data will help overcome concept drift, because the newer data better reflects the mapping from $X$ to $Y$ after concept drift.  $currentParams.rSize$ should never be less than $size(X)$, because we don't want to forget any new incoming data. $currentParams.rSize$ is computed in algorithm \ref{AdaptiverSize} below and will will be explained in subsection \ref{AdaptiveParameter}. 

\end{itemize}
\ \\ After the retained old data is concatenated with the incoming batch data, the decision tree is updated in a top-down fashion using the $UpdateSubTree(.)$ function based on $G(.)$.

\ \\In the $BuildSubTree(.)$ function, $minG(\cdot)$ will sort the data at the current node by all its features and store the order in the current node. By contrast, in the Incremental Update ($UpdateSubTree(.)$ function), $minGInc(\cdot)$ will sort only the incoming data (from $X$ and $Y$) and merge it with the previous data that is already sorted. Merging is of course faster than sorting, which conveys a time advantage to $UpdateSubTree(.)$ compared with $BuildSubTree(.)$.

\ \\At every interior node, function $minGInc(.)$ calculates a score for every feature by evaluating function $G$ on the data allocated to the current node. This calculation leads to the identification of the best feature and best value (or potentially values) to split on, while the splitting gives rise to two or more ranges for a feature. The data discarded in line 2 of (Algorithm \ref{DT:UpdateSubTree}) will not be considered by $minGInc(.)$. If, at some node, the  best splitting value (or values) is different from the choice before the arrival of the new data, the algorithm rebuilds the  subtree with the data retained in $current$ as well as the new data allocated to this node (the $BuildSubTree(.)$ function described in (Algorithm \ref{DT:BuildSubTree}). Otherwise, if the new best value range is the same as the range from the old tree, the algorithm splits only the incoming data among the children and then recursively calls the $UpdateSubTree(.)$ function on these children nodes. 

\ \\In summary, the forgetting strategy ensures that the model is  trained only on the newest $currentParams.rSize$ data. The rebuilding strategy determines whether a split point can be retained in which case tree reconstruction is vastly accelerated. Even if not,  the calculation of the split point based on $G(.)$ (e.g. Gini score) is somewhat accelerated because the relevant data is already nearly sorted.

\subsection{Adaptive Calculation of Retain Size and Max Tree Height}
\ \\ Retaining more historical data (larger $currentParams.rSize$)  will result in higher accuracy when there is no concept drift, because the old information is useful. When concept drift occurs, $currentParams.rSize$ should be small, because old information won't reflect the new concept (which is  some new mapping from input to label). Of course, a smaller $currentParams.rSize$ will result in increased speed. Thus, changing $currentParams.rSize$ can improve accuracy and reduce time. We use the following rules:

\begin{itemize}
     
\item When accuracy increases (i.e., the more recent predictions have been more accurate than previous ones) a lot,  the model can make good use of  more data, we want  $currentParams.rSize$ to increase with the effect that we discard little or no data.   When the accuracy increase is mild, the model has perhaps achieved an accuracy plateau, so we increase $currentParams.rSize$, but only slightly. 

\item When accuracy decreases, we want to decrease $currentParams.rSize$ to forget old data, because this indicates that concept drift has happened.  When accuracy decreases a lot, the new data may follow completely different rules from the old data, so we want to forget most of the old data, suggesting $currentParams.rSize$ should be very small. When concept drift is mild and accuracy decreases only a little, we want to retain more old data, so $currentParams.rSize$ should decrease only a little.

\item When accuracy changes little or not at all, we allow $currentParams.rSize$ to slowly increase. 
\end{itemize}

\label{AdaptiveParameter}
    \begin{algorithm}[H]
    \SetAlgoLined
	\SetKwInOut{Input}{input}
        \SetKwInOut{Output}{output}
        \SetKwInOut{Global}{global}
	\caption{AdaptParameters}
	\label{AdaptiverSize}
	\Input{ $\mb{X}$, $\mb{Y}$, incoming training data}
    \Output{
        $maxHeight$, the maximum height of the decision tree.
    }
    \Global{
	$currentParams.iRate$, the increase rate of $currentParams.rSize$ when the accuracy is stable
	\\ $currentParams.rSize$, the retain size of data
	\\ $currentParams.warmSize$, minimum size of the data for the decision tree to be considered ready to test
	\\ $currentParams.coldStartup$, $True$ if the decision tree is in Cold Startup mode.
        \\ $currentParams.(lastAcc, lastSize)$, the accuracy and data size of testing of previous update
        }
	
	\Begin{
        $guessAcc \leftarrow 1/nClasses$\;
        // $nClasses$ is the number of classes in $Y$\\
        $newAcc \leftarrow evaluate(\mb{X}, \mb{Y})-guessAcc$\;
        $lastAcc \leftarrow currentParams.lastAcc$\;
        // accuracy on latest batch\\
        \uIf{$currentParams.coldStartup==True$}{
        \While{$currentParams.rSize+size(X)>=currentParams.warmSize$}{\label{leaveColdCondition}
                $currentParams.warmSize \leftarrow 2*currentParams.warmSize$\;\label{selfAdjustWarmSize}
                \If{on the last 50\% of the data, accuracy is better than current $guessAcc$}{
                    $currentParams.coldStartup \leftarrow False$\; // no longer in startup mode
                    $currentParams.(lastAcc, lastSize) \leftarrow newAcc-1/nClasses, size(\mb{X})$\;
                    continue\;
            }\label{leaveCold}
        }
            $currentParams.rSize \leftarrow currentParams.rSize+size(X)$\; \label{coldStartUp}
        }
        \Else{
        \uIf{$newAcc<=0$} {\label{AdaptiverSize:special1} // we have left startup mode but our accuracy is bad, indicating concept drift
            $currentParams.rSize \leftarrow size(\mb{X})$
        }
        \uElseIf{$oldAcc<=0$}{
            $currentParams.rSize \leftarrow currentParams.rSize+size(X)$\;
        }\label{AdaptiverSize:special2}
        \Else{
            $currentParams.iRate \leftarrow currentParams.iRate*lastAcc/newAcc$\;\label{AdaptiverSize:iRate}
            $rChange \leftarrow (newAcc/lastAcc)^{(max(2, (3-newAcc/lastAcc)))}$\;
            $currentParams.rSize \leftarrow min(currentParams.rSize*rChange+currentParams.iRate*size(\mb{X}), currentParams.rSize+size(\mb{X})$\;\label{AdaptiverSize:min}
            $currentParams.(lastAcc, lastSize) \leftarrow newAcc, size(\mb{X})$\;
        }
        }
        \If{$currentParams.rSize < size(X)$}{
            $currentParams.rSize \leftarrow size(X)$\;
        }
        $maxHeight \leftarrow log2(currentParams.rSize)$\;
        }
    \end{algorithm}

\ \\To achieve the above requirements, we use the $AdaptParameters(.)$ function (Algorithm \ref{AdaptiverSize}). That function performs adaptively changes  $currentParams.rSize$, $maxHeight$, and $currentParams.iRate$ based on the change in accuracy. It is called when new data is acquired from the data stream and before function $UpdateSubTree(.)$ [Algorithm \ref{DT:UpdateSubTree}] is called. The $currentParams.rSize$ and $maxHeight$  will be applied to the parameters when calling $UpdateSubTree(.)$. The $currentParams.iRate$ and $currentParams.rSize$  will also be inputs to the next call to the $AdaptParameters(.)$ function on this tree. 


\ \\The $AdaptParameters(.)$ function will first test the accuracy of the model on new incoming data yielding $newAcc$. The function then recalls the accuracy that was tested last time as $lastAcc$. Next, because we want $newAcc$ and $lastAcc$ to improve upon random guessing, we subtract the accuracy of random guessing from $newAcc$ and from $lastAcc$ (the $guessAcc$ was already subtracted from $currentParams.lastAcc$ in the last update). We take the accuracy of random guessing ($guessAcc$)  to be $1/nClasses$.\footnote{We have tried to set $guessAcc$ to the accuracy of guessing the most frequent class always. That would be a good random strategy for skewed data. But after testing, we found $guessAcc=1/nClasses$ has a higher accuracy when the batch size is small and the dataset is very skewed towards one class. Otherwise, the two values of $guessAcc$ have similar accuracy.}  The intuitive reason to subtract $guessAcc$ is that a $lastAcc$ that is no greater than $guessAcc$ suggests that the model is no better than guessing just based on the number of classes. That in turn suggests that concept drift has likely occurred and so old data should be discarded. 

\ \\ Following that, $AdaptParameters(.)$ will calculate the rate of change ($rChange$) of $currentParams.rSize$ by the equation on line \ref{AdaptiverSize:iRate} of Algorithm \ref{AdaptiverSize}:
\begin{itemize}

    \item When $newAcc/lastAcc >= 1$, the $max$ in the exponent will ensure that $rChange$ will be $(newAcc/lastAcc)^{2}$. In this way, the $rChange$ curves slightly upward when $newAcc$ is equal to, or slightly higher than $lastAcc$, but curves steeply upward when $newAcc$ is much larger than $lastAcc$.
    
    \item When $newAcc/lastAcc < 1$, $rChange$ is equal to  $(newAcc/lastAcc)^{3-newAcc/lastAcc}$. In this way,  $rChange$ is flat or curves slightly downward when $newAcc$ is slightly lower than $lastAcc$ but curves very steeply downwards when $newAcc$ is much lower than $lastAcc$.
    
\end{itemize}

\ \\ Other functions to set $rChange$ are possible, but this one has the following properties that we expected: (i) it is continuous regardless of the values of newAcc and lastAcc; (ii) $rChange$ is close to 1 when $newAcc$ is close to $lastAcc$; (iii) when $newAcc$ differs from $lastAcc$ significantly in either direction, $rChange$ reacts strongly. 

\ \\Finally, we will calculate and update the new $currentParams.rSize$ by multiplying the old $currentParams.rSize$ by $rChange$. To effect a slow increase in $currentParams.rSize$ when $newAcc \approx lastAcc$ and $rChange \approx 1$, we increase $currentParams.rSize$ by $currentParams.iRate*size(X)$ in additional to $currentParams.rSize(old)*rChange$, where  $currentParams.iRate$ (the increase rate) is a number that is maintained from one call to $AdaptParameters(.)$ to another. The $min$ in line \ref{AdaptiverSize:min} reflects the fact that the size of the incoming data is $size(X)$, so $currentParams.rSize$ cannot increase by more than $size(X)$. Also, we do not allow $currentParams.rSize$ to be less than $size(X)$, because we do not want to forget any new incoming data. When $AdaptParameters(.)$  is called the first time, we will set $currentParams.rSize=size(X)+size(current.X)$.

\ \\  The two special cases of lines \ref{AdaptiverSize:special1} through \ref{AdaptiverSize:special2} happen when $newAcc<=0$ or $lastAcc<=0$.  When $newAcc<=0$, the prediction of the model is no better than random guessing. In that case, we infer that the old data cannot help in predicting new data, so we will forget all of the old data by setting $currentParams.rSize=size(X)$. When $lastAcc<=0$ but $newAcc>0$, then all the old data may be useful. So we set $currentParams.rSize=currentParams.rSize+size(X)$. 
 

\ \\The above adaptation strategy requires a dampening parameter $currentParams.iRate$ to limit the increase rate of $currentParams.rSize$. When the accuracy is large, the model may be close to its maximum possible accuracy, so we may want a smaller $currentParams.iRate$ and in turn to increase $currentParams.rSize$ slower. After a drastic concept drift event, when the accuracy has been significantly decreased, we want to increase $currentParams.iRate$ to retain more new data after forgetting most of the old data. This will shorten the duration of the cold start after the concept drift. 
To achieve this, we will adaptively change it as follows: $currentParams.iRate(new)=currentParams.iRate(old)*lastAcc/newAcc$ before each time $currentParams.rSize$ is updated, as in line \ref{AdaptiverSize:iRate} of (algorithm \ref{AdaptiverSize}). $currentParams.iRate$ will not be changed if either $lastAcc<=0$ or $newAcc<=0$.

\ \\ Upon initialization, if the first increment is small, then  $newAcc$ may not exceed $1/nClasses$, and the model will forget all of the old data every time. To avoid such poor performance at cold start, the Forgetful Decision Tree will be initialized in Cold Startup mode. In that Cold Startup mode, the Forgetful Decision Tree will not forget any data (line \ref{coldStartUp} in algorithm \ref{AdaptiverSize}). When  $currentParams.rSize$ reaches $currentParams.warmSize$, the Forgetful Decision Tree will leave Cold Startup mode if $newAcc$ is better than $guessAcc$ since the last $50\%$ of data arrived. Otherwise, $currentParams.warmSize$ will be doubled. The above process will be repeated until leaving the Cold Startup mode. 

\ \\ Max tree height is closely related to the size of data retained in the tree. We want each leaf node to have about one data item on average when the tree is perfectly balanced, so we always set $maxHeight=log2(currentParams.rSize)$. 

\ \\ In summary, algorithm \ref{AdaptiverSize} provide a method to adjust $currentParams.rSize$, $maxHeight$, and $currentParams.iRate$. We still need an initial value of $currentParams.iRate$. We show experimentally how to find that in section \ref{Tune IR}.


\subsection{Forgetful random forests.}
\label{frf}

\ \\ When new data is acquired from the data stream, the $UpdateForest(.)$ function (Algorithm \ref{RF:Update}) will incrementally update the entire random forest. 

\ \\The Forgetful Random Forest (Algorithm \ref{RF:Update}) is based on the Forgetful Decision Tree described above in section \ref{forgetdecisiontrees}. Each Random Forest contains $nTree$ decision trees. The $UpdateSubTreeRF(.)$ and $BuildSubTreeRF(.)$ functions for each decision tree in the Random Forest are the same as those from the decision tree algorithms described in section \ref{forgetdecisiontrees} except:

\begin{itemize}
    \item Only a limited number of features are considered at each split, increasing the chance of invoking $UpdateSubTree(.)$ during recursion, and thus saving time by avoiding the need to rebuild subtrees from scratch. The number of features considered by each Random Forest tree is uniformly and randomly chosen within range $(\lfloor\sqrt{nFeatures}\rfloor+1, nFeatures]$, where $nFeatures$ is the number of features in the dataset. The features considered by each ensemble tree are randomly and uniformly selected without replacement when created and will not change in subsequent updates. Further, every node inside the same ensemble tree considers the same features.
    \item To increase the diversity in the forest, $UpdateSubTreeRF(.)$ function will randomly and uniformly discard old data without replacement, instead of discarding data based on time of insertion.
    \item To decrease the correlation between trees and increase the diversity in the forest, we give the user the option to choose the  Leveraging Bagging \cite{10.1007/978-3-642-15880-3_15} strategy to the data arriving at each Random Forest tree. The size of the data after bagging is $W$ times the size of original data, where $W$ is a random number with a expected value of $6$, generated by a $Poisson(\lambda=6)$ distribution. To avoid the performance hit  resulting from  too many copies of the data, we  never allow $W$ to be larger than $10$. Each data item in  the expanded data is randomly and uniformly selected from the original data with replacement. We  apply bagging to each decision tree inside the random forest.
\end{itemize}

\subsection{Discard Poorly Performing Trees}

\ \\To discard the trees with features that perform poorly after concept drift, we will  call the $Discard(.)$ function (Algorithm \ref{RF:Discard}) when $newAcc$ is significantly less than $lastAcc$. As in algorithm \ref{AdaptiverSize}, we will subtract the accuracy of random guessing $guessAcc$ from $newAcc$ and $oldAcc$ to show the improvement of the model with respect to random guessing. Significance is based on a $p-value$ test: the accuracy of the forest has changed with a $p-value<tThresh$ based on a 2 sample t-test. The variable $tThresh$ is a hyper-parameter that will  be tuned in section \ref{Tune tThresh}. 

\ \\To detect slight but continuous decreases in accuracy, we will update $lastAcc$ and $lastSize$ by averaging them with $newAcc$ and $size (X)$ when we observe an insignificant change in accuracy. In other cases, when the change in accuracy is significant, we will replace $lastAcc$ and $lastSize$ with $newAcc$ and $size (X)$ after the $Discard(.)$ function is called.

\ \\The $Discard(.)$ function (Algorithm \ref{RF:Discard}) removes  $((newAcc-lastAcc)/lastAcc)*nTree$ Random Forest trees having the least accuracy when evaluated on the new data. The discarded trees are replaced with new decision trees. Each new tree will take all the data from the tree it replaced, but the tree will be rebuilt, the $currentParams$ for that tree will be re-initialized, and the considered features will be re-selected for that tree. After building the new tree, the algorithm will test the tree on the latest data to calculate  $nTree.lastAcc$ and $newTree.lastSize$. In this way, new trees  adapt their $currentParams.rSize$ and $currentParams.iRate$ with the newly arriving data.

\begin{algorithm}[H]
    \SetAlgoLined
	\SetKwInOut{Input}{input}
	\SetKwInOut{Output}{output}
	\SetKwInOut{Global}{global}
	\caption{UpdateForest}\label{RF:Update}
	\Input{ ($\mb{X}$, $\mb{Y}$), incoming training data for current node
        \\$nTree$, the number of decision trees that are actively updated
        \\$firstCall$, whether UpdateForest is called for the first time
        \\$tThresh$, the threshold for discarding trees
	\\ $E(\cdot)$, the stopping criteria, returns boolean 
        \\ $G(\cdot)$, a function to score the fitness of a feature for splitting}
        \Global{
        $current.trees$ the list of decision trees that are actively updated
        \\$current.(lastAcc, lastSize)$ the accuracy and data size of testing of of previous update
        \\$current.allCurrentParams$, the parameters that will be updated for each tree in $current.trees$
        }
	\Begin{
        $newAcc \leftarrow fractionCorrect(current, X, Y)-1/nClasses$\;
        // $nClasses$ is the number of classes in $Y$\\
        $lastAcc \leftarrow current.lastAcc$\;
        \uIf{$2Sample\_t\_test(newAcc, lastAcc, size(X), current.lastSize)>tThresh$}{
            \If{$newAcc<lastAcc$}{
                $current.trees, current.allCurrentParams \leftarrow Discard(((newAcc-lastAcc)/lastAcc)*nTree, \mb{X}, \mb{Y}, current.trees, current.allCurrentParams, E(.), G(.))$;
            }
            $current.lastAcc \leftarrow newAcc$\;
            $current.lastSize \leftarrow  size(\mb{X})$\;
        }\Else{
            $current.lastAcc \leftarrow \frac{(lastAcc*current.lastSize+newAcc*size(X))}{current.lastSize+size(X)}$\;
            // lastAcc is updated based on a weighted average, which is weighted by the size of the new data compared to the previous data.
            \\$current.lastSize \leftarrow  current.lastSize+size(X)$\;
        }
        
        \uIf{$firstCall$}{
            \For{each $tree$ in $current.trees$}{
                $maxHeight \leftarrow log2(size(X))$\;
                $BuildSubTreeRF(Bagging(\mb{X}, \mb{Y}), tree, E(maxHeight), G(.))$\;
            }
        }\Else{
            \For{each $tree$ in $current.trees$}{
                $baggedX, baggedY \leftarrow Bagging(X, Y)$\;
                $maxHeight, current.allCurrentParams[tree] \leftarrow AdaptParameters(baggedX, baggedY, current.allCurrentParams[tree])$\;
                $tree \leftarrow UpdateSubTreeRF(baggedX, baggedY, tree, E(maxHeight), G(.)$, $tree.(X, Y, gRecord, ranges, label),  current.allCurrentParams[tree].rSize)$\;
            }
        }
    }
\end{algorithm}

\begin{algorithm}[H]
    \SetAlgoLined
	\SetKwInOut{Input}{input}
	\SetKwInOut{Global}{global}
	\caption{Discard}\label{RF:Discard}
	\Input{ 
	  $nDiscard$, the number of trees that will be discarded
	\\($\mb{X}$, $\mb{Y}$), incoming training data for current node
	\\ $E(\cdot)$, the stopping criteria, returns boolean 
        \\ $G(\cdot)$, a function to score the fitness of a feature for splitting}
        \Global{
            $current.trees$, the list of decision trees that are actively updated
            \\$current.allCurrentParams$, the parameters that will be updated for each tree in $current.trees$
        }
	\Begin{
        $discardTree \leftarrow$ $nDiscard$ trees in $current.trees$ having the smallest accuracy (fractionCorrect)\;
        \For{each $tree$ in $discardTree$}{
            $treeNew \leftarrow$ create a new Tree\;
            $treeNew \leftarrow BuildSubTreeRF(tree.X, tree.Y, treeNew, E(.), G(,)$\;
            Initialize $current.allCurrentParams[treeNew]$\;
            $current.allCurrentParams[treeNew] \leftarrow$ $evaluate(treeNew, tree.X, tree.Y)$\;
            replace tree with treeNew in $current.trees$\;
        }
    }
\end{algorithm}



\section{Tuning the Values of the Hyperparameters}\label{hyperparameters}

\ \\Decision trees and random forests have six hyperparameters to set, which are $G(\cdot)$, $E(\cdot)$, $currentParams.warmSize$, $tThresh$, $nTree$, and $currentParams.iRate$. To find the best values for these hyperparameters, we generated 18 datasets with different intensity of concept drifts, number of concept drifts, and Gaussian noise, using the generator inspired by \cite{Synthetic}.

\ \\After testing on the 18 new datasets, we find that some hyperparameters have optimal values (with respect to accuracy) that apply to all datasets. Others have values that vary depending on the dataset but can be learned. 


\ \\Here are the ones whose optimal values can be fixed in advance.

\begin{itemize}
\item
The evaluation function $G(\cdot)$ can be a Gini Impurity\cite{gini} score or an entropy reduction coefficient. Which one is chosen doesn't make a material difference, so we set $G(\cdot)$ to entropy reduction coefficient for all datasets.
\item
The maximum tree height ($maxHeight$) is adaptively set based on the methods in section \ref{AdaptiveParameter} to log base 2 of $currentParams.rSize$. Applying other stopping criteria does not materially affect the accuracy. For that reason, we ignore other stopping criteria.
\item
To mitigate the inaccuracies of a cold startup, the model will not discard any data in Cold Startup mode. To leave Cold Startup mode, we will test the accuracy is better than random guessing on the last $50\%$ of data, when $currentParams.rSize$ is at least $currentParams.warmSize$. $currentParams.warmSize$ adapts if it is too small (line \ref{selfAdjustWarmSize} in algorithm \ref{AdaptiverSize}), so we will set its initial value to $64$ data items.
\end{itemize}

\ \\The parameters that must be learned through training are:
\begin{itemize}
\item
$currentParams.iRate$, the rate of increase of $currentParams.rSetain$, the retained data,  when the accuracy changes little. As shown in algorithm \ref{AdaptiverSize}, it adapts over time. But it starts at some initial value which needs to be set.
\item
$tThresh$ is the threshold of $p-value$ to discard inaccurate trees inside a Forgetful Random Forest. A small $tThresh$ will result in always discarding trees even without concept drift, but a large $tThresh$ will result in never discarding trees even with concept drift.
\item
$nTree$ is the number of trees inside the forest. Larger $nTree$ will result in better accuracy up to a point.
\end{itemize}







    \begin{figure}
        \centering
        \caption{$currentParams.iRate$ against Accuracy: $currentParams.iRate=0.3$ usually results in a high accuracy}
        \label{fig:TuningG}
        \includegraphics[width=0.45\textwidth]{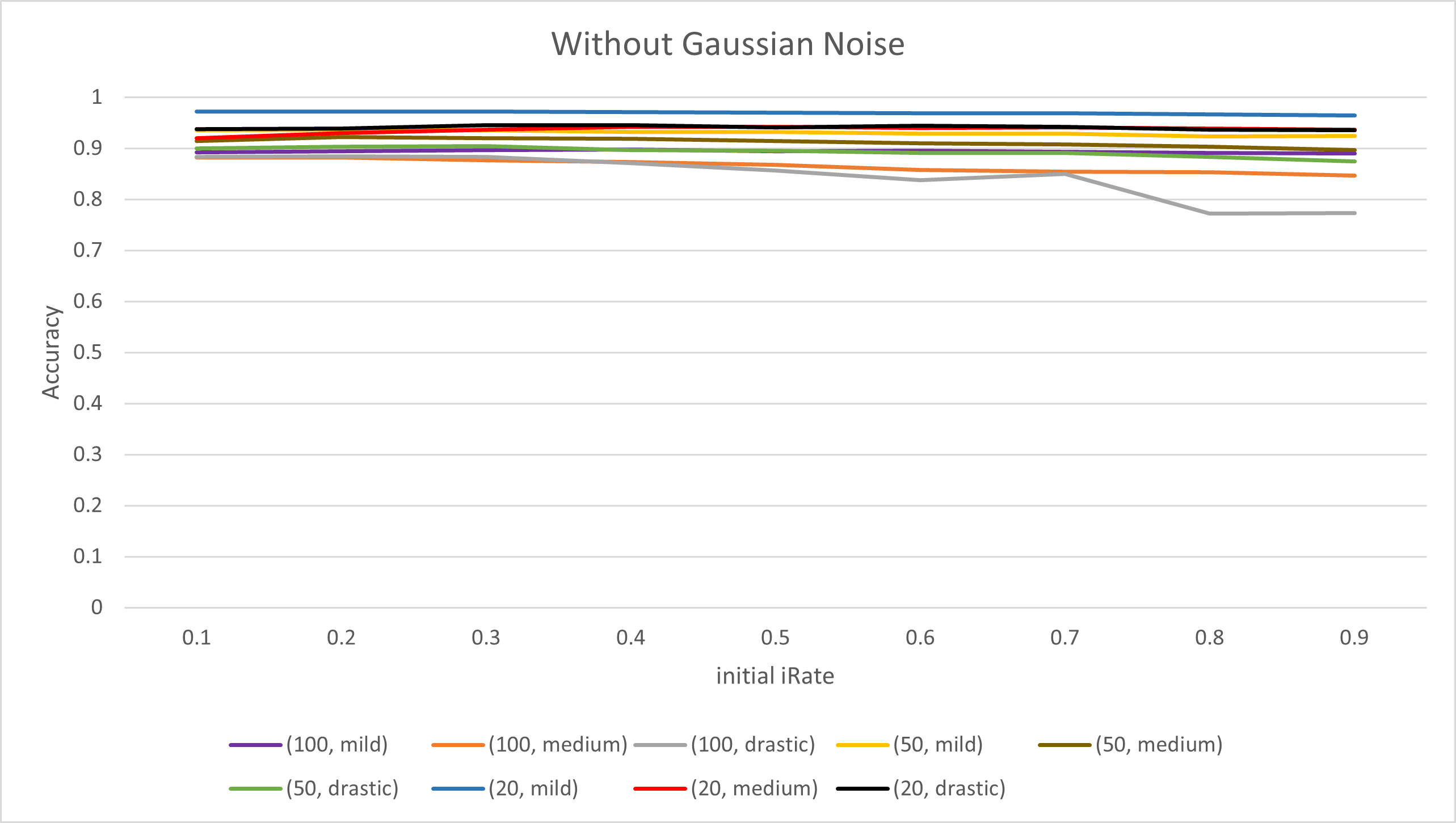}
        \includegraphics[width=0.45\textwidth]{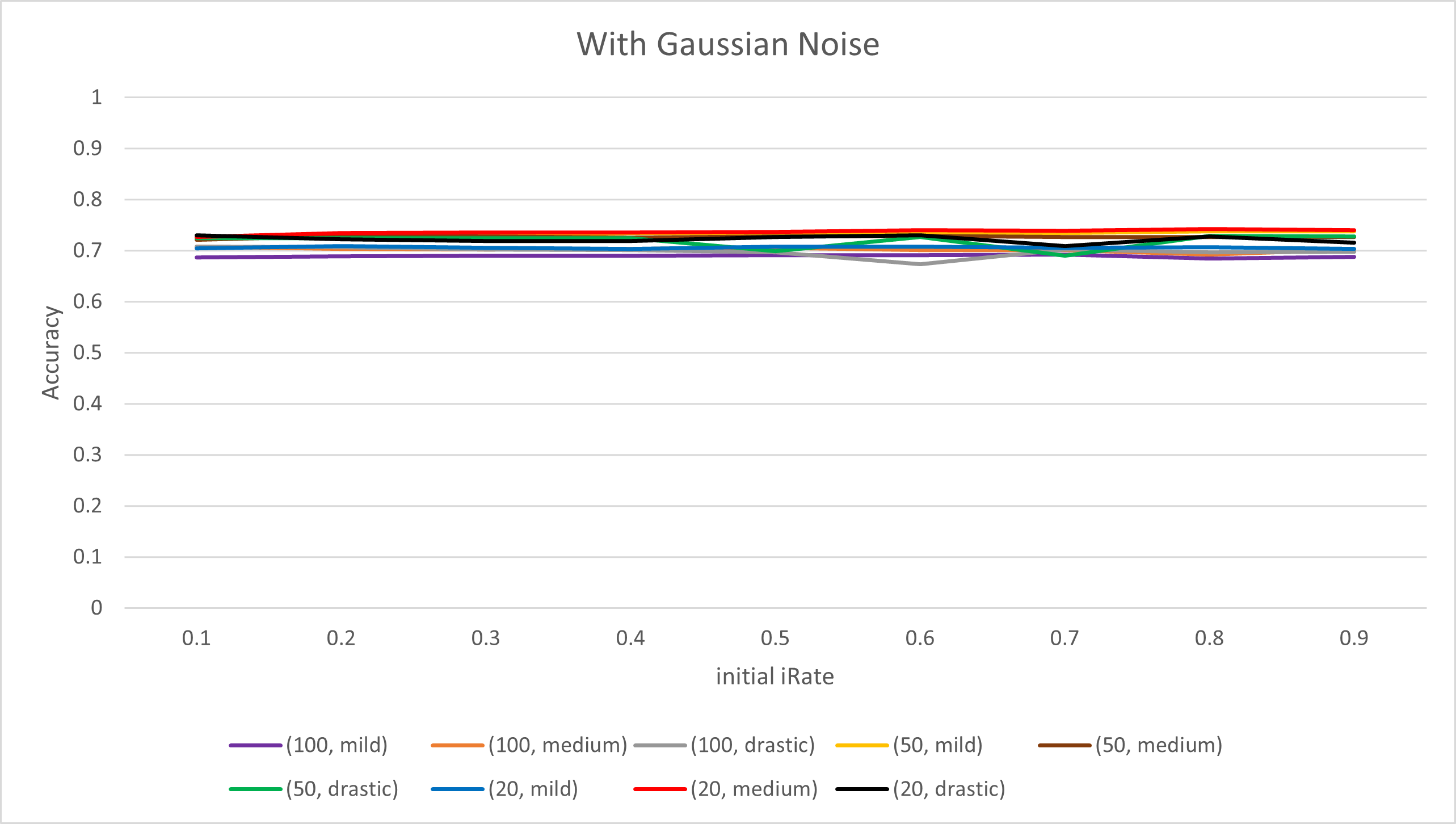}
    \end{figure}
    
    \begin{figure}
        \centering
        \caption{$currentParams.iRate$ after each Update: $currentParams.iRate$  changes significantly from the initial value, which is 0.3, over the course of the executions.}
        \label{fig:iRate}
        \includegraphics[width=0.45\textwidth]{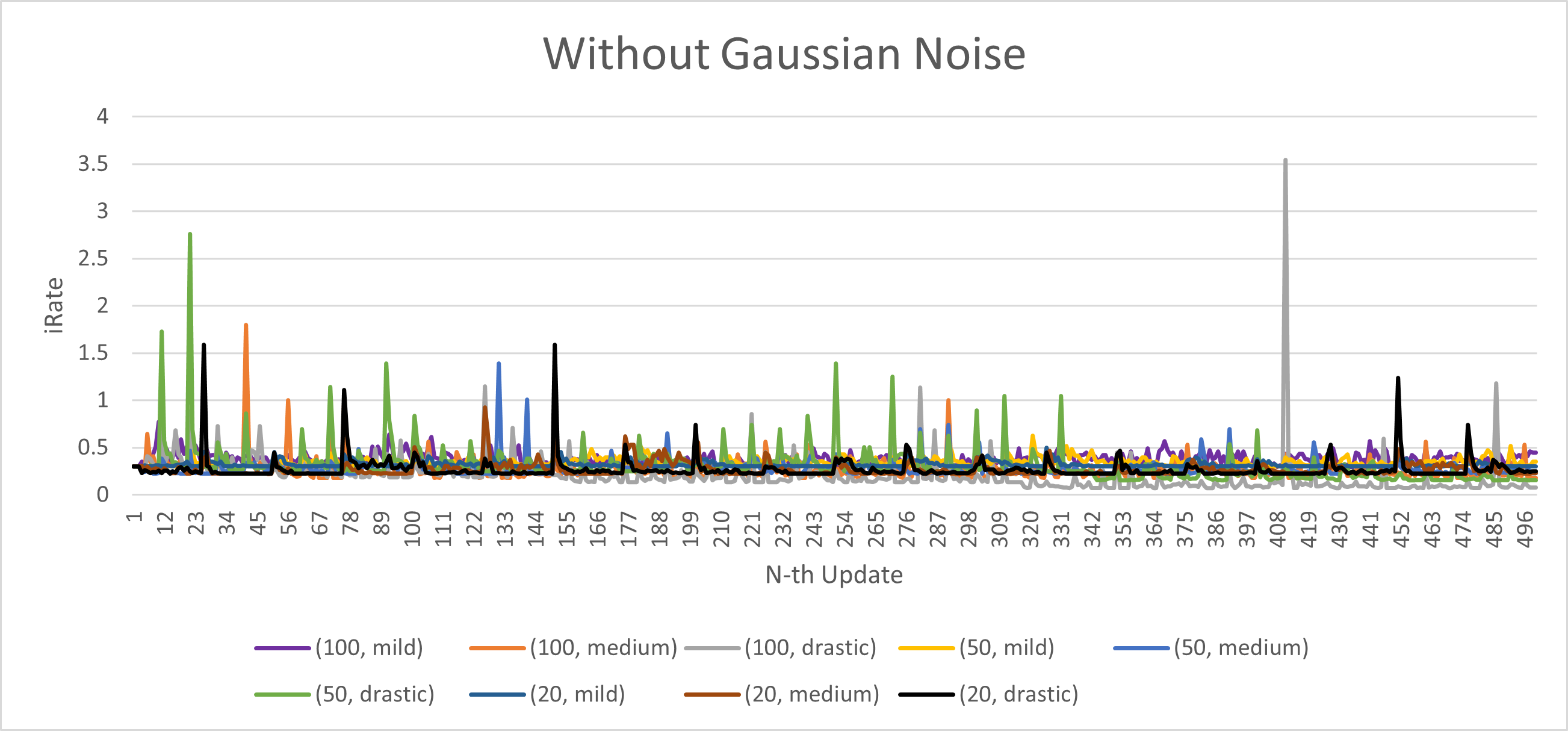}
        \includegraphics[width=0.45\textwidth]{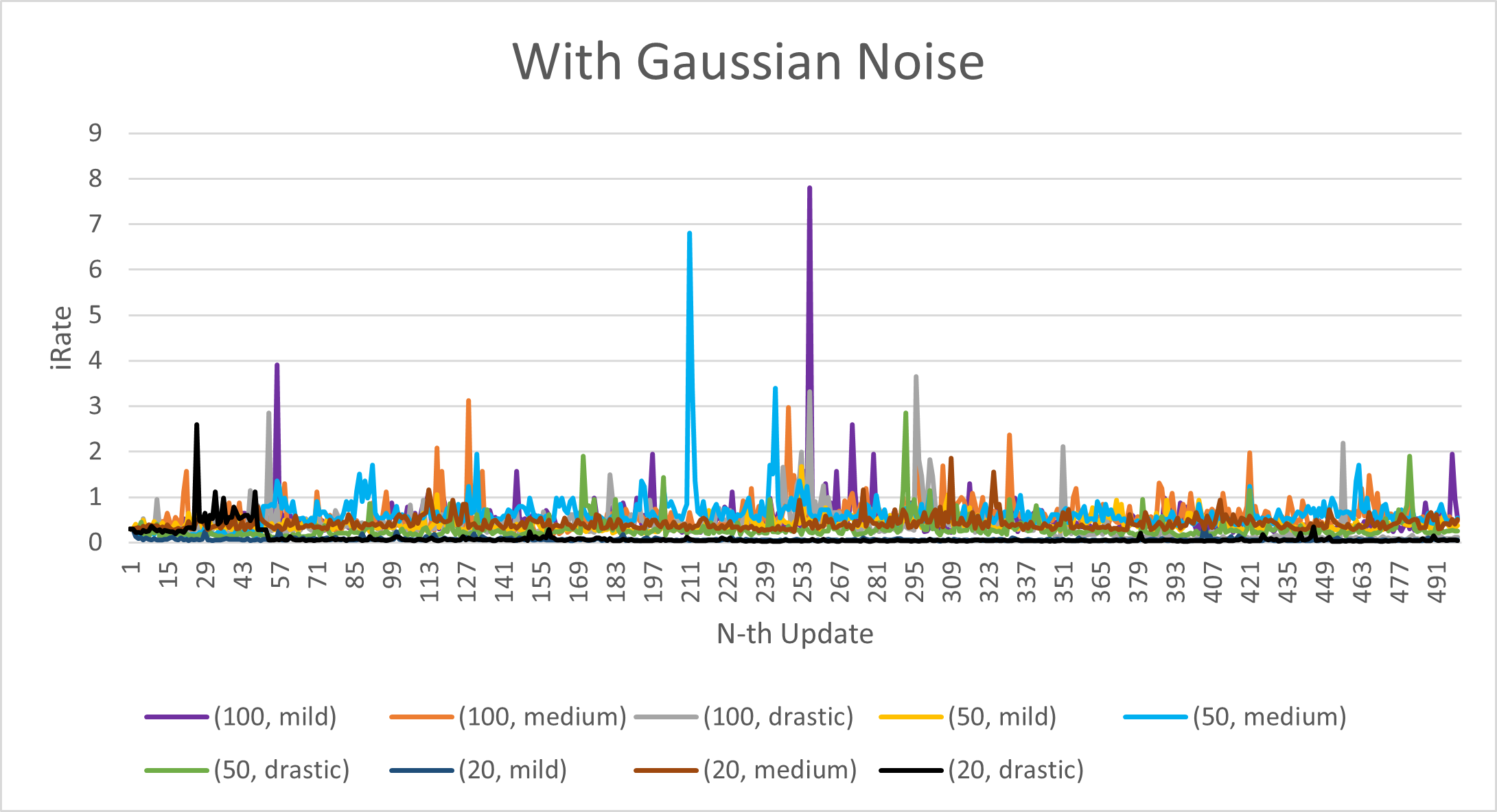}
    \end{figure}
    
\subsection{Tuning Increase Rate}
\label{Tune IR}

\ \\The adaptation strategy in section \ref{AdaptiveParameter} need an initial value for parameter $currentParams.iRate$ which influences the increase rate of $currentParams.rSize$. Too much data will be retained if the initial $currentParams.iRate$ is large, but cold start will last too long if $currentParams.iRate$ is too small. 

\ \\To find the best initial $currentParams.iRate$, we created 18 simulated datasets. Each dataset contains $50,000$ data items labeled with ${0, 1}$ without noise. Each item is characterized by 10 binary features. Each dataset is labeled with $(C, I)$, where $C$ means that it has $(C-1)$ uniformly distributed concept drifts, and $I$ is the intensity of the concept drift, while a mild concept drift will drift one feature, a medium concept drift will drift 3 features, and a drastic concept drift will drift 5 features. Also, for each dataset, we have one version without Gaussian Noise, and the other version with Gaussian Noise ($\lambda =0$, $std=1$, unit is the number of features). We tested the Forgetful Decision Tree with different initial $currentParams.iRate$ values on these synthetic datasets, while other hyperparameter values were fixed in advance as in section \ref{hyperparameters}.


\ \\From Figure \ref{fig:TuningG}, we observe that the Foregetful Decision Tree does well when $currentParams.iRate=0.3$ or $currentParams.iRate=0.4$ initially. We will use $currentParams.iRate=0.3$ as our initial setting and use it in the experiments of section \ref{experiment} for all our algorithms, because most simulated datasets have higher accuracy at $currentParams.iRate=0.3$ than at $currentParams.iRate=0.4$.

\ \\Also, from Figure \ref{fig:iRate}, we observe a significant increase of $currentParams.iRate$ after each concept drift. This implies that new data accumulates after concept drift.  However, the hyperparameter $currentParams.iRate$ will decrease to a low level after the accuracy starts to increase, thus avoiding the retention of too much data.

\subsection{Tuning Discard Threshold }
\label{Tune tThresh}

    \begin{figure}
        \centering
        \caption{$tThresh$ against Accuracy without Bagging: $tThresh=0.05$ usually results in a high accuracy}
        \label{fig:TuningtThresh}
        \includegraphics[width=0.45\textwidth]{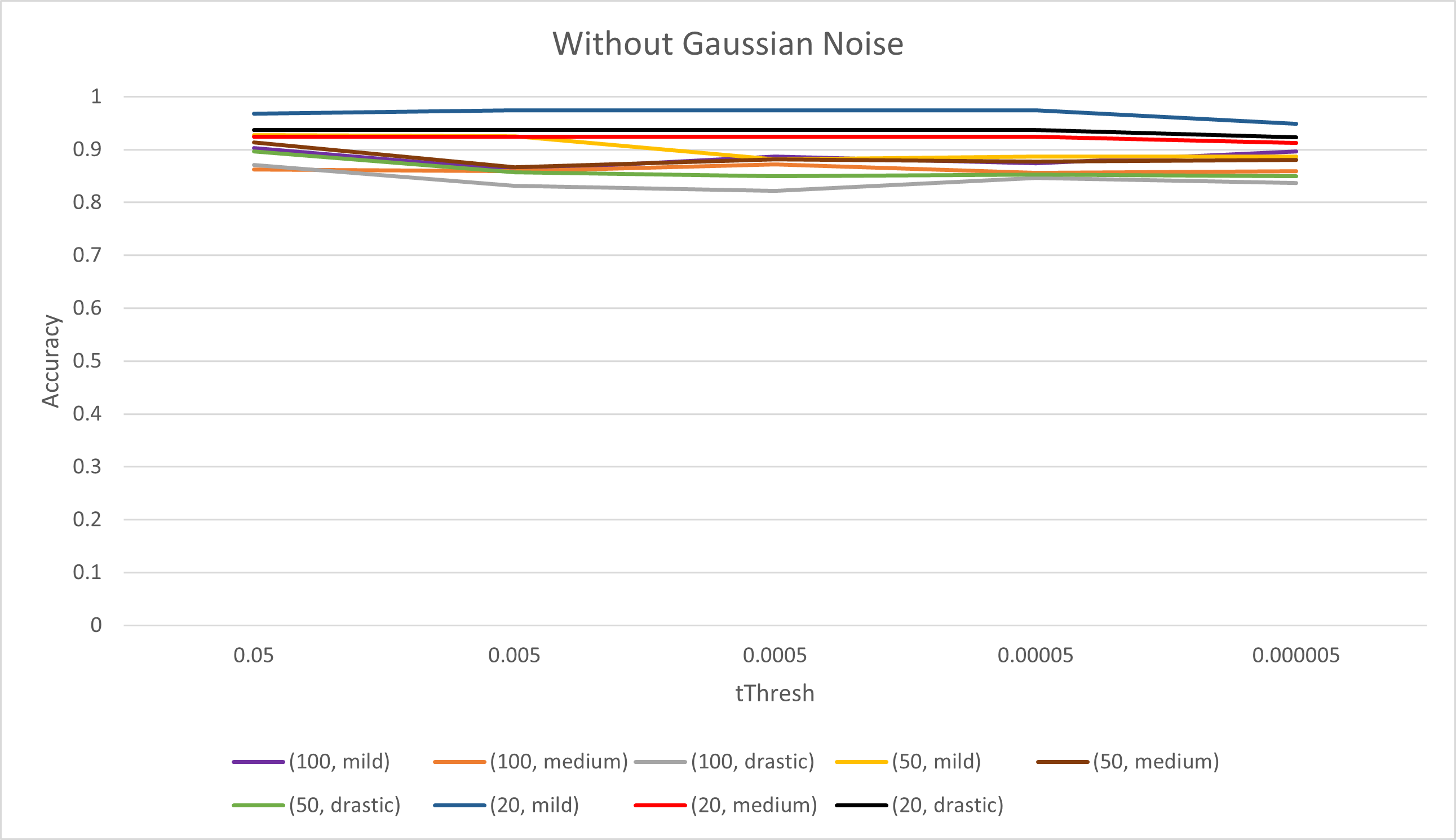}
        \includegraphics[width=0.45\textwidth]{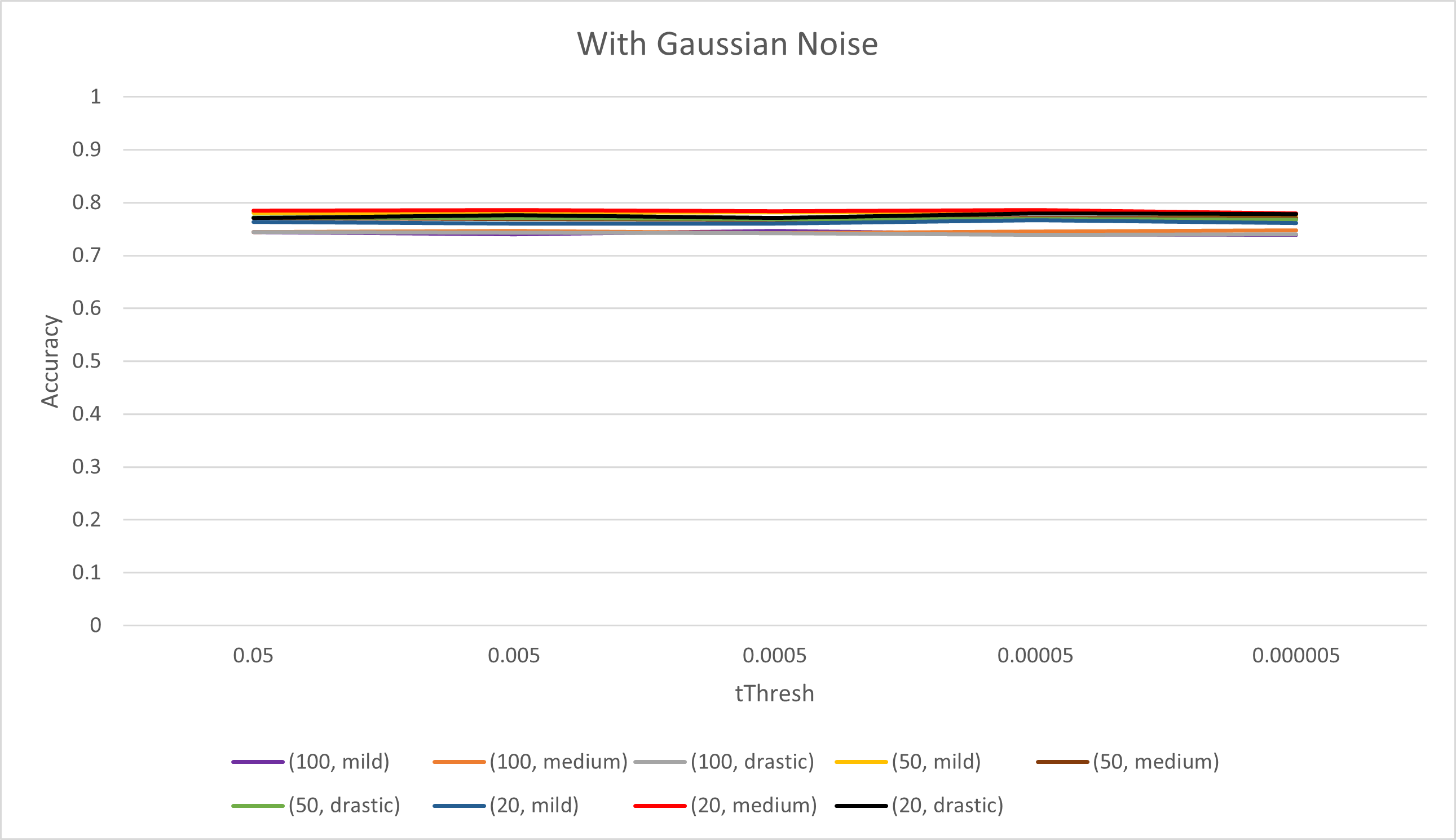}
    \end{figure}

    \begin{figure}
        \centering
        \caption{$tThresh$ against Accuracy with Bagging: $tThresh=0.05$ usually results in a high accuracy}
        \label{fig:TuningtThreshBag}
        \includegraphics[width=0.45\textwidth]{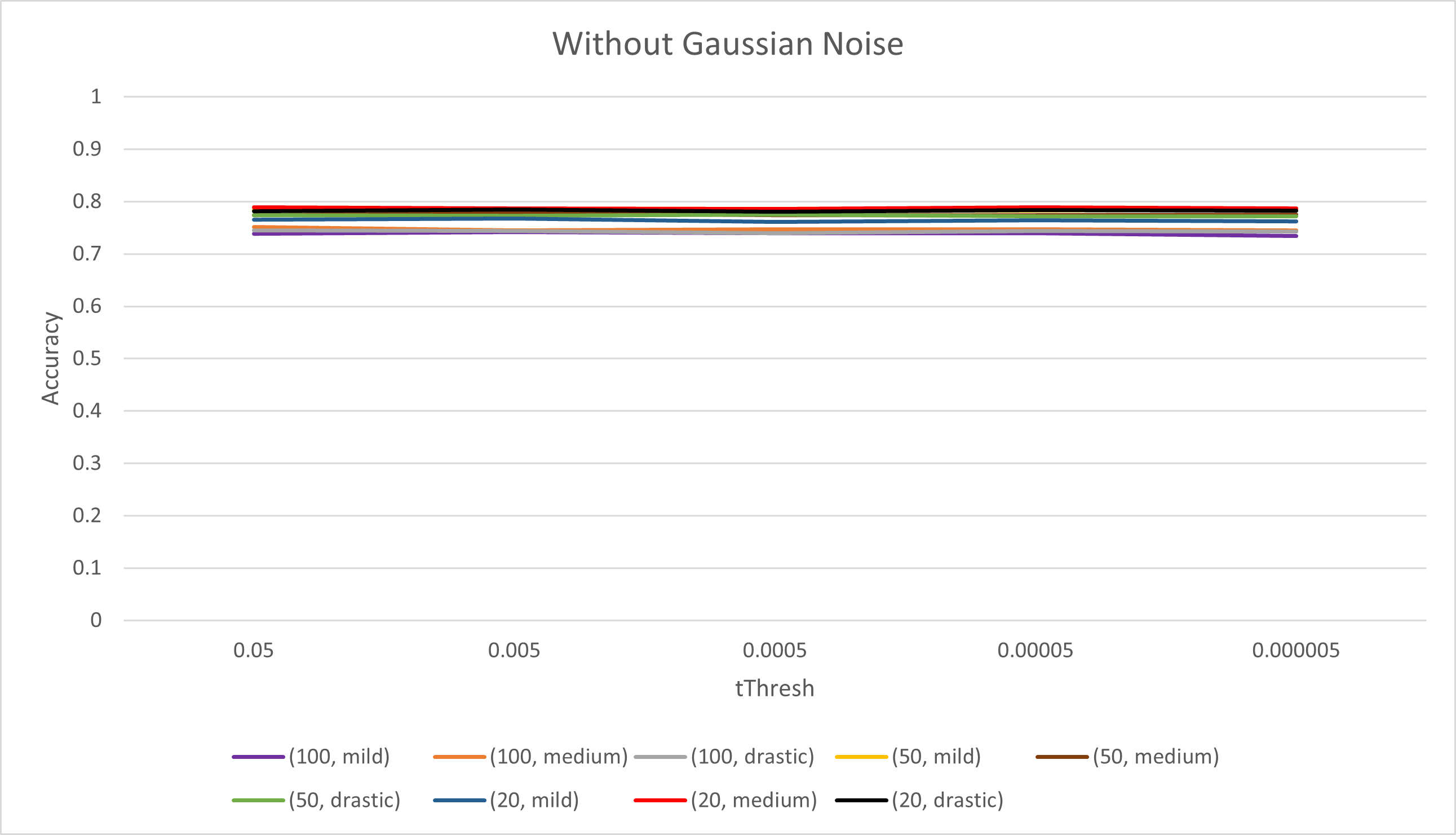}
        \includegraphics[width=0.45\textwidth]{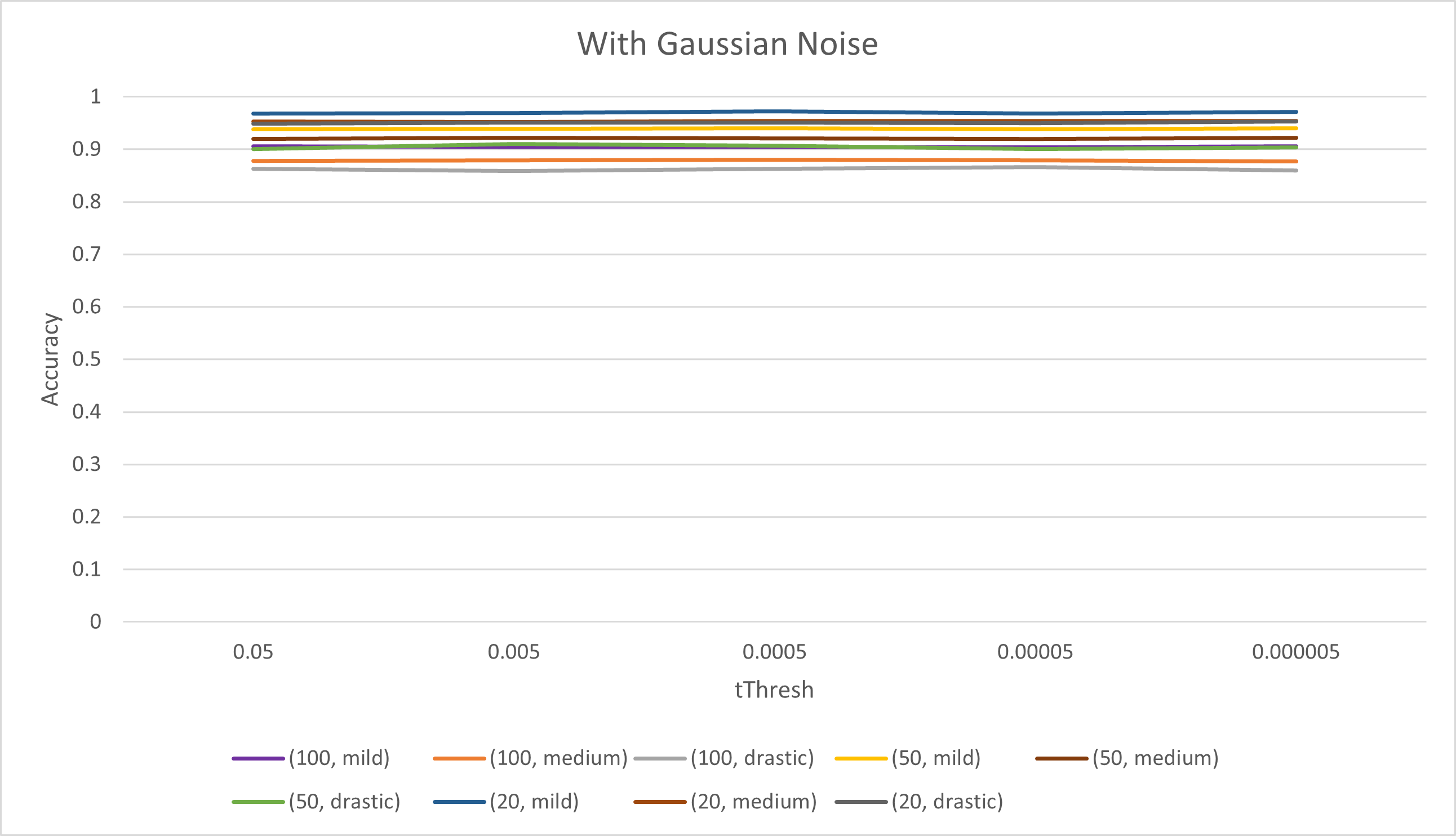}
    \end{figure}

\ \\To find the best $tThresh$, we use the same simulated datasets as in section \ref{Tune IR}. We tested the Forgetful Random Forest with different $tThresh$ values on these synthetic datasets, with other hyperparameter values fixed in advance as in section \ref{hyperparameters}. From Figure \ref{fig:TuningtThresh} and Figure \ref{fig:TuningtThreshBag}, we observe that all datasets enjoy a good accuracy when $tThresh=0.05$, for both with bagging and without bagging.

\subsection{Tuning Number of Trees}
\label{tunenTree}
    \begin{figure}
        \centering
        \caption{$nTree$ against Accuracy without bagging: The accuracy stops increasing after $nTree>=20$}
        \label{fig:TuningnTree}
        \includegraphics[width=0.45\textwidth]{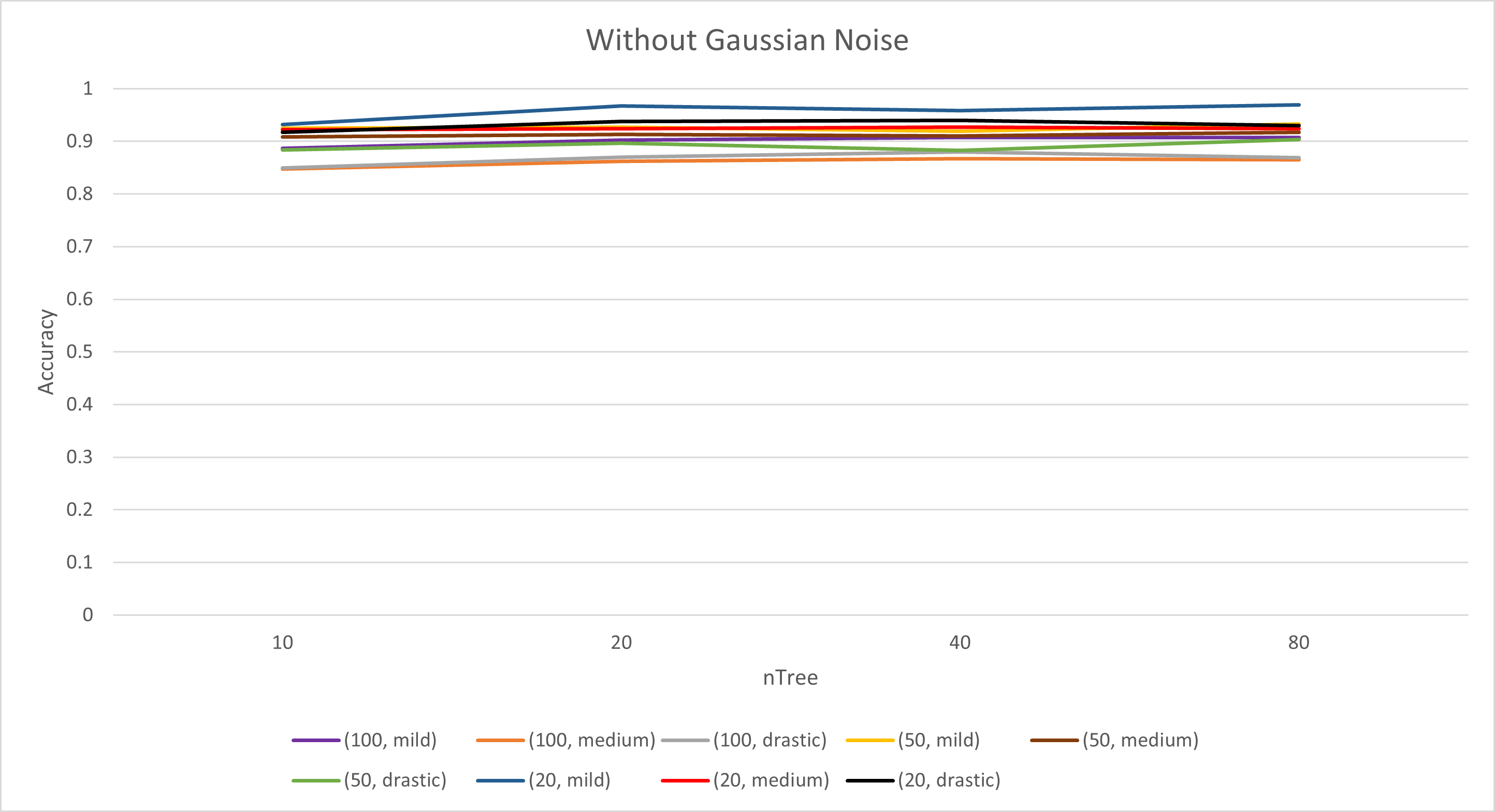}
        \includegraphics[width=0.45\textwidth]{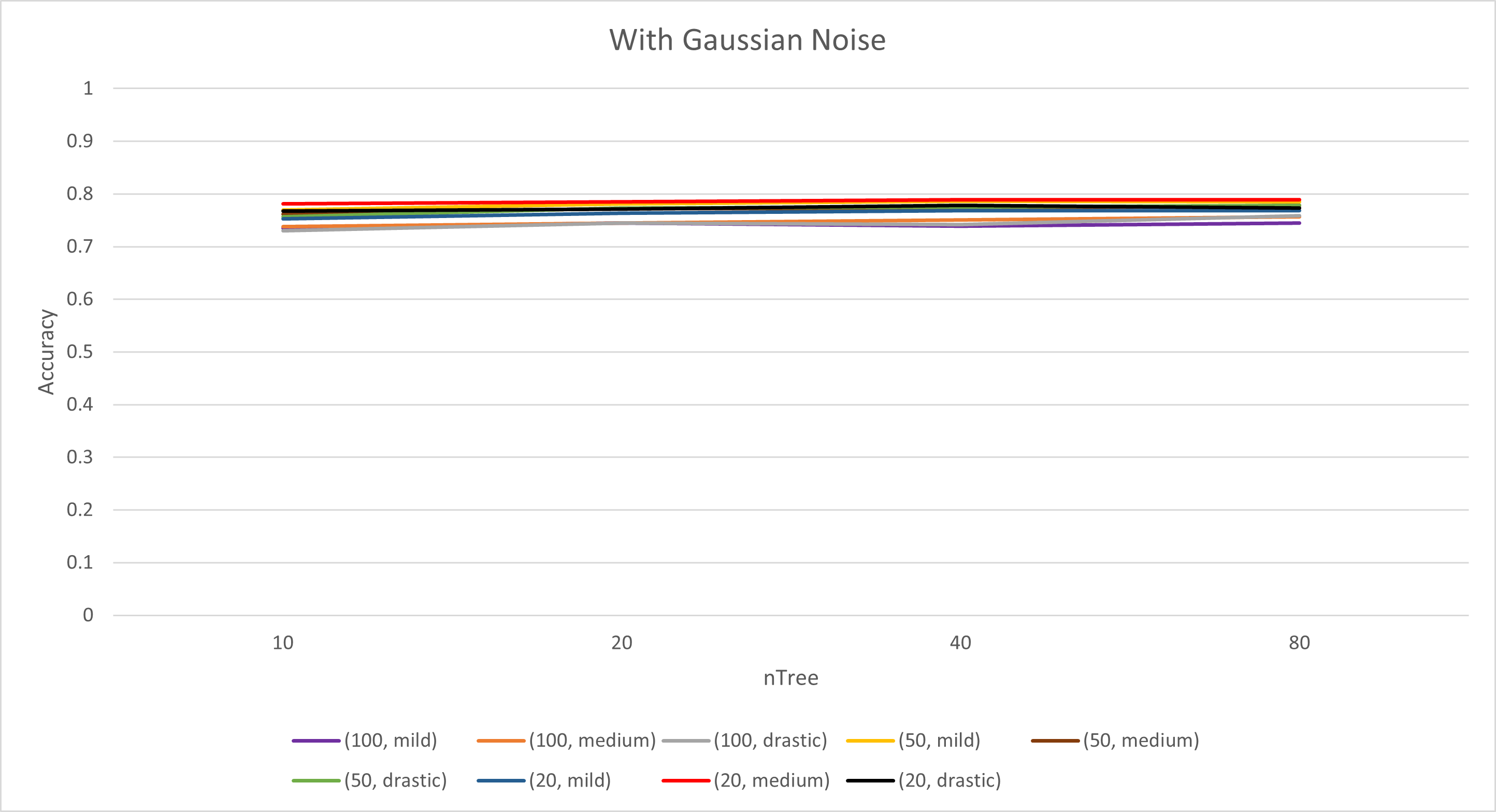}
    \end{figure}
    \begin{figure}
        \centering
        \caption{$nTree$ against Accuracy with bagging: The accuracy stops increasing after $nTree>=20$}
        \label{fig:TuningnTreeBag}
        \includegraphics[width=0.45\textwidth]{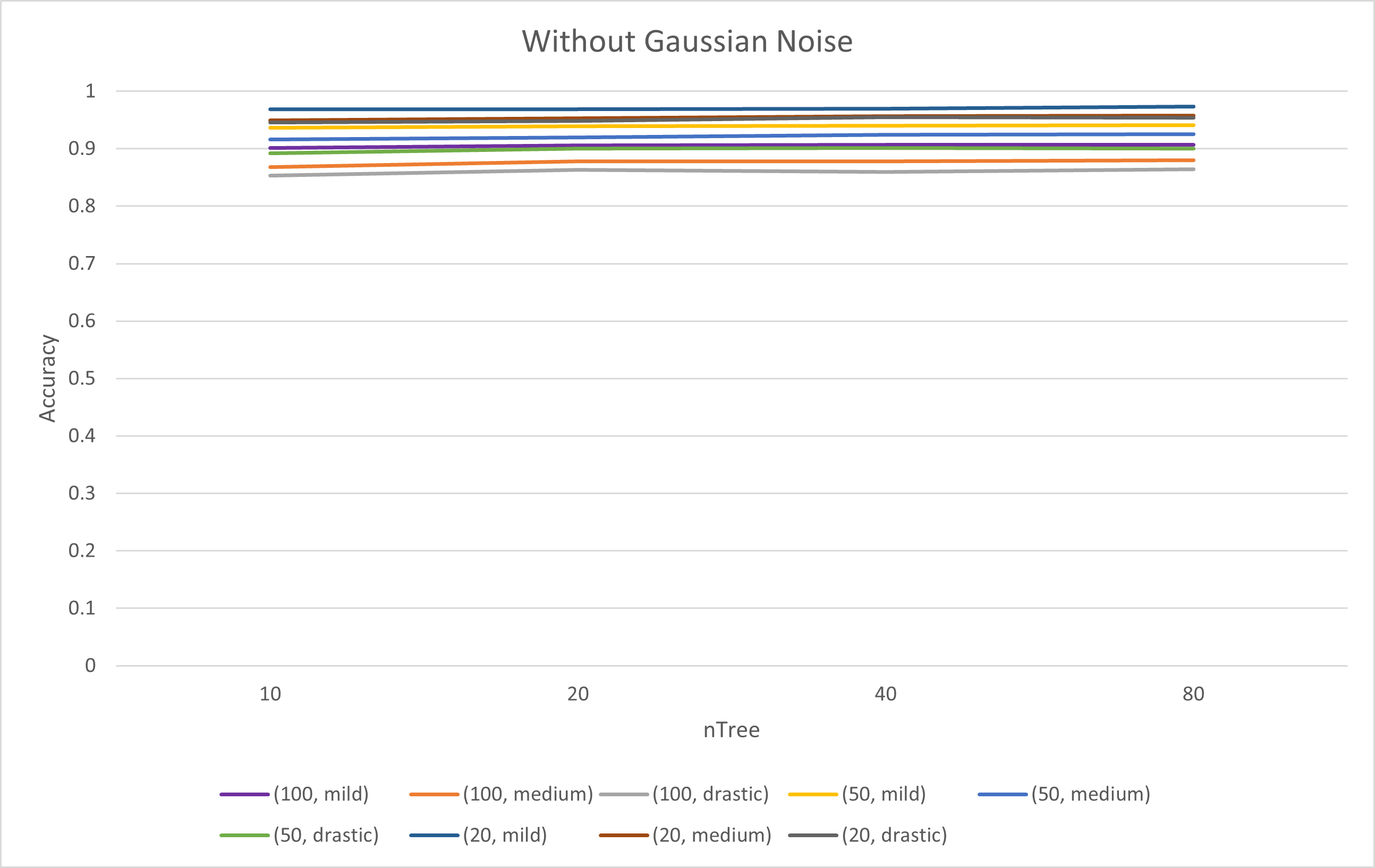}
        \includegraphics[width=0.45\textwidth]{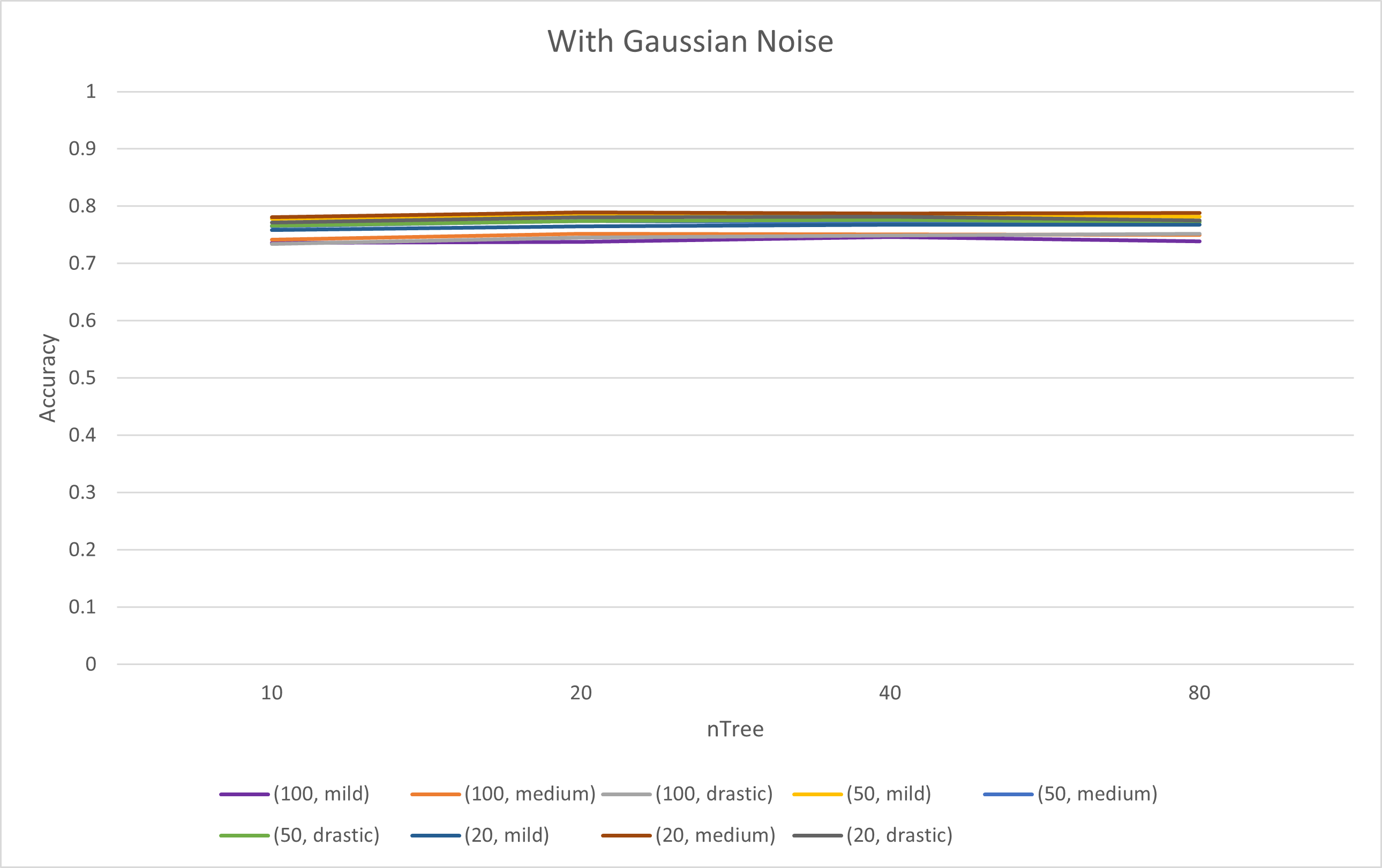}
    \end{figure}

\ \\To find the best $nTree$ value, we use the same simulated datasets as in section \ref{Tune IR}. We tested Forgetful Random Forest with different $nTree$ values on these synthetic datasets, while the other hyperparameter values were fixed in advance as in section \ref{hyperparameters}. From Figure \ref{fig:TuningnTree} and Figure \ref{fig:TuningnTreeBag}, we observe that the accuracy of all datasets stops growing after $nTree>20$ for both with bagging and without bagging, so we will set $nTree=20$.

\subsection{Main Decision Tree Algorithm}
\label{MainDT}

The Forgetful Decision Tree main routine ($ForgetfulDecisionTree$) is called initially and then each time a new batch (an incremental batch) of data is received. The routine will make predictions with the tree before the batch and then update the batch. The algorithm \ref{algorithm:DTMain} describes only the update part. The prediction part is what would be done by any Decision Tree. Accuracy results are recorded only after the accuracy flattens out, which means the accuracy changes 10\% or less between the last 500 data items and the previous 500 data items.

\begin{algorithm}[H]
    \SetAlgoLined
	\SetKwInOut{Input}{input}
	\caption{ForgetfulDecisionTree}\label{algorithm:DTMain}
	\Input{ 
	  \\ $dataStream$, the stream of data}
	\Begin{
            $E(.) \leftarrow MaximumTreeHeight(.)$ \;
            $G(.) \leftarrow Entropy(.)$\;
            $currentParams.iRate \leftarrow 0.3$\;
            $currentParams.coldStartup \leftarrow True$\;
            $currentParams.warmSize \leftarrow$ 64 data items\;
            $currentParams.(lastAcc, lastSize) \leftarrow (0, 0)$\;
            $root \leftarrow$ new decision tree root node\;
            $initialFlag \leftarrow True$\;
            \While{ receiving new batch $X, Y$ from $dataStream$}{
                \uIf{$initialFlag$}{
                    $initialFlag \leftarrow False$\;
                    $currentParams.rSize \leftarrow size(X)$\;
                    $maxHeight \leftarrow log2(currentParams.rSize)$\;
                    $root \leftarrow BuildSubTree(X, Y, root, E(maxHeight), G(.))$\;
                }\Else{
                    $maxHeight, currentParams \leftarrow AdaptParameters(X, Y, currentParams.(iRate, rSize, warmSize, coldStartup, lastAcc, lastSize)$\;
                    $root \leftarrow UpdateSubTree(X, Y, root, E(maxHeight), G(.), root.(X, Y, gRecord, ranges, label), currentParams.rSize))$
                }
            }
        }
\end{algorithm}

\subsection{Main Random Forest Algorithm}
\label{MainRF}

The Forgetful Random Forest main routine ($ForgetfulRandomForest$) is called initially and then each time an incremental batch of data is received. The routine will make predictions with the Random Forest before the batch and then update the Random Forest. The algorithm \ref{algorithm:RFMain} describes only the update part. The prediction part is what would be done by any Random Forest. Accuracy results will apply after the accuracy flattens out, which means the accuracy changes 10\% or less between the last 500 data items and the previous 500 data items.


\begin{algorithm}[H]
    \SetAlgoLined
	\SetKwInOut{Input}{input}
	\caption{ForgetfulRandomForest}\label{algorithm:RFMain}
	\Input{ 
	  $features$, all features in data
	  \\ $dataStream$, the stream of data for training}
	\Begin{
            $E(.) \leftarrow MaximumTreeHeight(.)$ \;
            $G(.) \leftarrow Entropy(.)$\;
            $current.lastAcc \leftarrow 0$\;
            $current.lastSize \leftarrow 0$\;
            $current.trees \leftarrow$ $nTree$ new decision trees\;
            $nTree \leftarrow 20$\;
            $tThresh \leftarrow 0.05$
            \For{each $tree$ in $current.trees$}{
                //Initialize $consider$ of each $tree$\\
                $nConsider \leftarrow$ single value from uniform distribution random variable between $(\lfloor \sqrt{size(features)} \rfloor + 1$ and $ size(features)]$ \;
                $allConsider[tree] \leftarrow$ uniformly and randomly select $nConsider$ features from $features$ without replacement\; 
                $current.allCurrentParams[tree].iRate \leftarrow 0.3$\;
                $current.allCurrentParams[tree].coldStartup \leftarrow True$\;
                $current.allCurrentParams[tree].warmSize \leftarrow$ 64 data items\;
                $current.allCurrentParams[tree].(lastAcc, lastSize) \leftarrow (0, 0)$
            }
            $firstCall \leftarrow True$\;
            \While{ receiving new batch $X, Y$ from $dataStream$}{
                $firstCall \leftarrow False$\;
                $current \leftarrow UpdateForest(X, Y, nTree, firstCall, tThresh, current.(lastAcc, lastSize, trees), allConsider$, $current.allCurrentParams, E(.), G(.))$\;
            }
    }
\end{algorithm}

\section{Experiments}\label{experiment}
\ \\ This section compares the following algorithms: Forgetful Decision Tree, Forgetful Random Forest with bagging, Forgetful Random Forest without bagging, Hoeffding Tree \cite{VFDT} \cite{HT}, Hoeffding Adaptive Tree\cite{10.1007/978-3-642-03915-7_22}, iSOUP Tree\cite{Osojnik2017TreebasedMF}, and Adaptive Random Forest\cite{AdaptiveRF}. The forgetful algorithms  from section \ref{ForgetfulDataStructure} use the hyperparameter settings from section \ref{hyperparameters} on both real  and simulated datasets produced by others. 

We measure time consumption, the accuracy and, where appropriate, the F1-score. 

Because we have tuned the forgetful data structures, we have also tuned the state-of-the-art algorithms on the same generated data sets of section \ref{hyperparameters}. 
The following settings yield the best accuracy for the state-of-the-art algorithms

\begin{itemize}
    \item Previous papers (\cite{VFDT} and \cite{HT}) provide two different configurations for the Hoeffding Tree. The configuration from \cite{VFDT} usually has the highest accuracy, so we will use it in the following experiment: $split\_confidence=10^{-7}$, $grace\_period=200$, and $tie\_threshold=0.05$. Because the traditional Hoeffding Tree cannot deal with concept drift, so we set $leaf\_prediction=Naive Bayes Adaptive$ to allow the model to adapt when concept drift happens.
    \item The designer of Hoeffding Adaptive Tree suggests six versions of configuration of the Hoeffding Adaptive Tree, which are HAT-INC, HATEWMA, HAT-ADWIN, HAT-INC NB, HATEWMA NB, and HAT-ADWIN NB. HAT-ADWIN NB has the best accuracy, and we will use it in the following experiment. The configuration is $leaf\_prediction=Naive Bayes$, $split\_confidence=0.0001$, $grace\_period=200$, and $tie\_threshold=0.05$.
    \item The designer provides only one configuration for iSOUP-Tree \cite{Osojnik2017TreebasedMF}, so we will use it in the following experiment. The configuration is $leaf\_prediction=adaptive$, $split\_confidence=0.0001$, $grace\_period=200$, and $tie\_threshold=0.05$. 
    \item The designer of Adaptive Random Forest provided six variant configurations of Adaptive Random Forest \cite{AdaptiveRF}: the variants $ARF_{moderate}$, $ARF_{fast}$, $ARF_{PHT}$, $ARF_{noBkg}$, $ARF_{stdRF}$, and $ARF_{maj}$.  $ARF_{fast}$ has the highest accuracy in most cases that we tested, so we will use that configuration: $\delta_w = 0.01$, $\delta_d = 0.001$, and $learners=100$. 
\end{itemize}

\subsection{Hyperparameter Settings for Forgetful Data Structures}
\ \\The previous sections gave us the following hyperparameters settings:
\begin{itemize}
\item 
$G(\dot)$ is entropy reduction for all of the Forgetful Random Forests and the Forgetful Decision Tree.
\item 
$currentParams.warmSize$ should be initially small, because it can be adaptively increased. We initially set it to $64$ data items.
\item
Following the tuning of section \ref{Tune IR}, we will initialize $currentParams.iRate$ to $0.3$.
\item
$tThresh$ defines the threshold for discarding trees inside a Random Forest. Following the tuning of section \ref{Tune tThresh}, we will set $tThresh=0.05$.
\item
$nTree$ is the number of trees inside the Forgetful Random Forest. Following the tuning of section \ref{tunenTree}, we set it to $20$ for both versions of the Forgetful Random Forests.
\end{itemize}

\subsection{Metrics}

\ \\Beside {\bf accuracy}, we also use {\bf precision}, {\bf recall}, and {\bf F1-score} to evaluate our methods. Precision and recall are appropriate to problems where there is a
class of interest and the question is which percentage of predictions of that class are correct (precision) and how many instances of that class are predicted (recall). This is appropriate for the phishing application where the question is whether the website is a phishing website. Accuracy is more appropriate in all other applications. For example, in the electricity datasets, price up and price down are both classes of interest. Therefore, we present precision, recall, and the F1-score for Phishing only. We use the following formula based on the confusion matrix for the following tests.
\begin{itemize}
    \item $accuracy = \frac{|True Positive| + |True Negative|}{Size(test-set)}$
    \item $precision = \frac{|True Positive|}{|True Positive|+|False Positive|}$
    \item $recall = \frac{|True Positive|}{|True Positive|+|False Negative|}$
    \item $F1-score = \frac{|2*True Positive|}{2*|True Positive|+|False Negative|+|False Positive|}$
\end{itemize}

\ \\In contrast to most static labeled machine learning tasks, we don't partition the data into a training set and a test set. Instead, when each batch of data arrives, we measure the accuracy and F1-score of the predictions on that incremental set, before we use it to update the models. We start measuring accuracy and F1-score after the accuracy of Forgetful Decision Tree flattens out, in order to avoid the inaccuracies during start-up. That point is different for each dataset as described in section \ref{datasets}, but all algorithms will start measuring the accuracy and F1-score at same point for the same dataset.


\subsection{Datasets}
\label{datasets}

\ \\Because the real datasets all contain categorical variables and our methods don't handle those directly, we modify the categorical variables into their one-hot encodings using the OneHotEncoder of sklearn \cite{JMLR:v12:pedregosa11a}. For example,  a categorical variable $color=\{R, G, B\}$ will be transfered to three binary variables $isR=\{True, False\}$, $isG=\{True, False\}$, and $isB=\{True, False\}$. Also, all of the forgetful methods in the following tests use only binary splits at each node.


\ \\ To measure statistical stability in the face of the noise caused by randomization caused by the setting of the initial seeds, we will test all decision trees and random forests six times with different seeds and record the average values with $95\%$ confidence interval for time consumption, accuracy, and F1-score.

\ \\ We use the following three real datasets and two synthetic datasets to test the performance of our forgetful methods against the state-of-the-art incremental algorithms.

\begin{itemize}
\item Forest Cover Type (ForestCover) \cite{covertype} dataset corresponds to the forest cover type for a given observation (30 x 30 meter cell) determined from the US Forest Service (USFS) Region 2 Resource Information System (RIS) data. Each increment consists of 400 observations. This dataset suffers concept drift because different ranges have different mappings from input image to forest type. For this dataset, the accuracy flattens out after the first $24,000$ data items have been observed, out of $581,102$ data items. Each increment arrives in the sequence from the original dataset. 
\item The Electricity \cite{Electricity} dataset describes the price and demand of electricity. The goal is to forecast the price trend in the next 30 minutes Each increment consists of data from one day. This data suffers from concept drift, because of market and other external influences. For this dataset, the accuracy keeps changing and never flattens out, so we will start measuring accuracy after the first increment, which is after the first $49$ data items have arrived, out of $36,407$ data items?
\item Phishing \cite{Phishing} contains contains 11,055 web pages accessed over time, some of which are malicious.  Each increment consists of 100 pages. The tactics of Phishing purveyors get more sophisticated over time, so this dataset suffers from concept drift. For this dataset, the accuracy flattens out after the first $500$ data items have arrived.
\item Two Synthetic datasets from \cite{Synthetic}.  Both are time-ordered and are designed to suffer from concept drift over time. One, called \ii{Gradual}, has $41,000$ data points. Gradual is characterized by complete label changes that happen gradually over $1,000$ data points at three single points, and $10,000$ data items between each concept drift. Another dataset, called \ii{Abrupt}, has $40,000$ data points. It undergoes complete label changes at three single points, and $10,000$ data items between each concept drift. Each increment consists of $100$ data points. Unlike the datasets that were used in section \ref{hyperparameters}, these datasets contains only four features, two of them are binary classes without noise, and the other two are sparse values generated by $sin(x)$ and $sin^{-1}(y)$, where $x$ and $y$ are the uniformly generated random numbers. 
For both of these datasets, the accuracy flattens out after $1000$ data items have arrived. 
\end{itemize}

The following experiments are performed with NYU HPC, with one core cpu and 200 GB memory. 
\ \\
\ \\
\subsection {Quality and Time Performance of Forgetful Decision Tree}
\label{exp:DT}

    \begin{figure}
        \centering
        \caption{Time Consumption of Decision Trees: Based on this logarithmic scale, the Forgetful Decision Tree is at least three times faster than the state-of-the-art incremental Decision Trees.}
        \label{fig:DT_T}
        \includegraphics[width=0.8\textwidth]{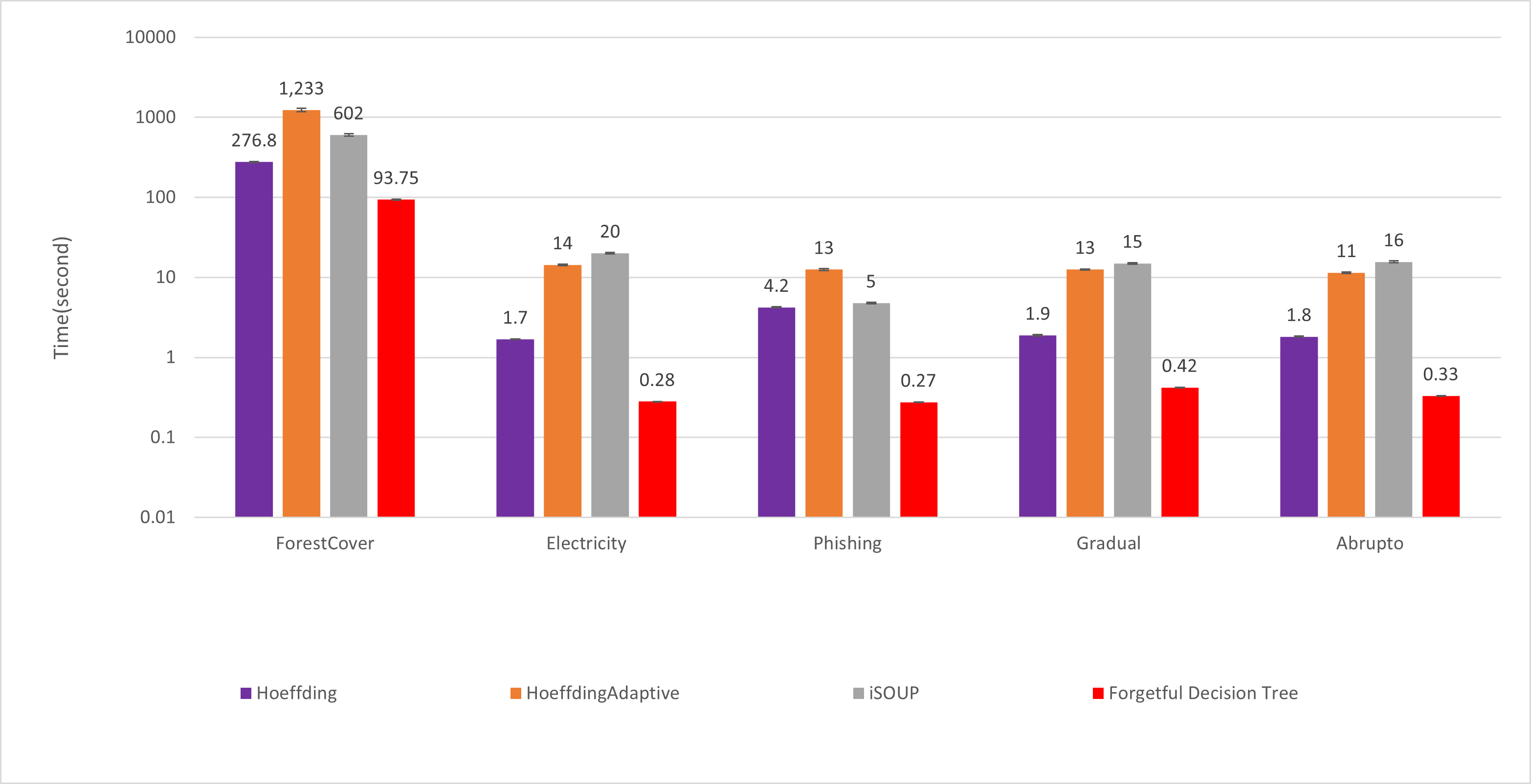}
    \end{figure}
    \begin{figure}
        \centering
        \caption{Accuracy of Decision Trees: The Forgetful Decision Tree is at least as accurate as the state-of-the-art incremental Decision Trees(iSOUP-tree) and at most $9\%$ more accurate.}
        \label{fig:DT_Acc}
        \includegraphics[width=0.8\textwidth]{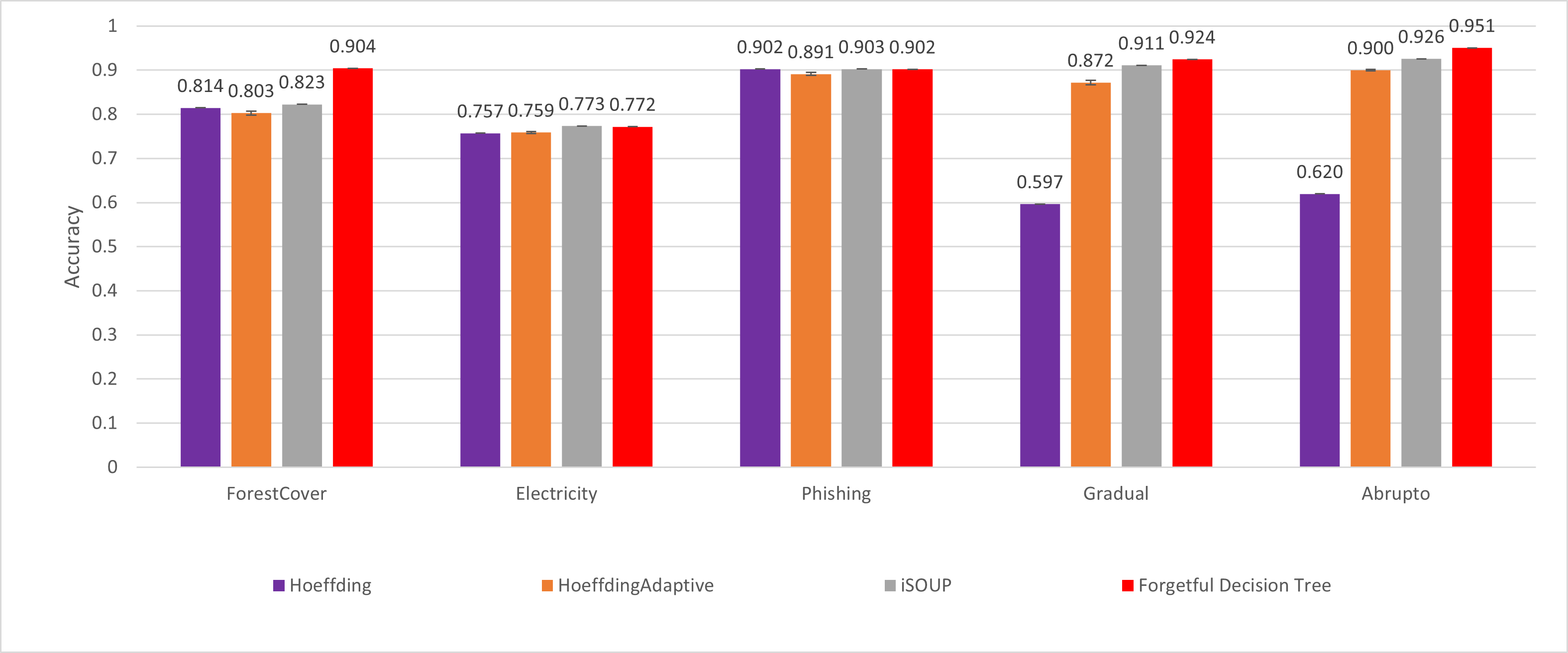}
    \end{figure}
    \begin{figure}
        \centering
        \caption{Precision, Recall, and F1-score of Decision Trees: While precision and recall results vary, the Forgetful Decision Tree has 
        a similar F1-score  to the other incremental Decision Trees for the Phishing dataset (the only one where F1-score is appropriate).}
        \label{fig:DT_F1}
        \includegraphics[width=0.9\textwidth]{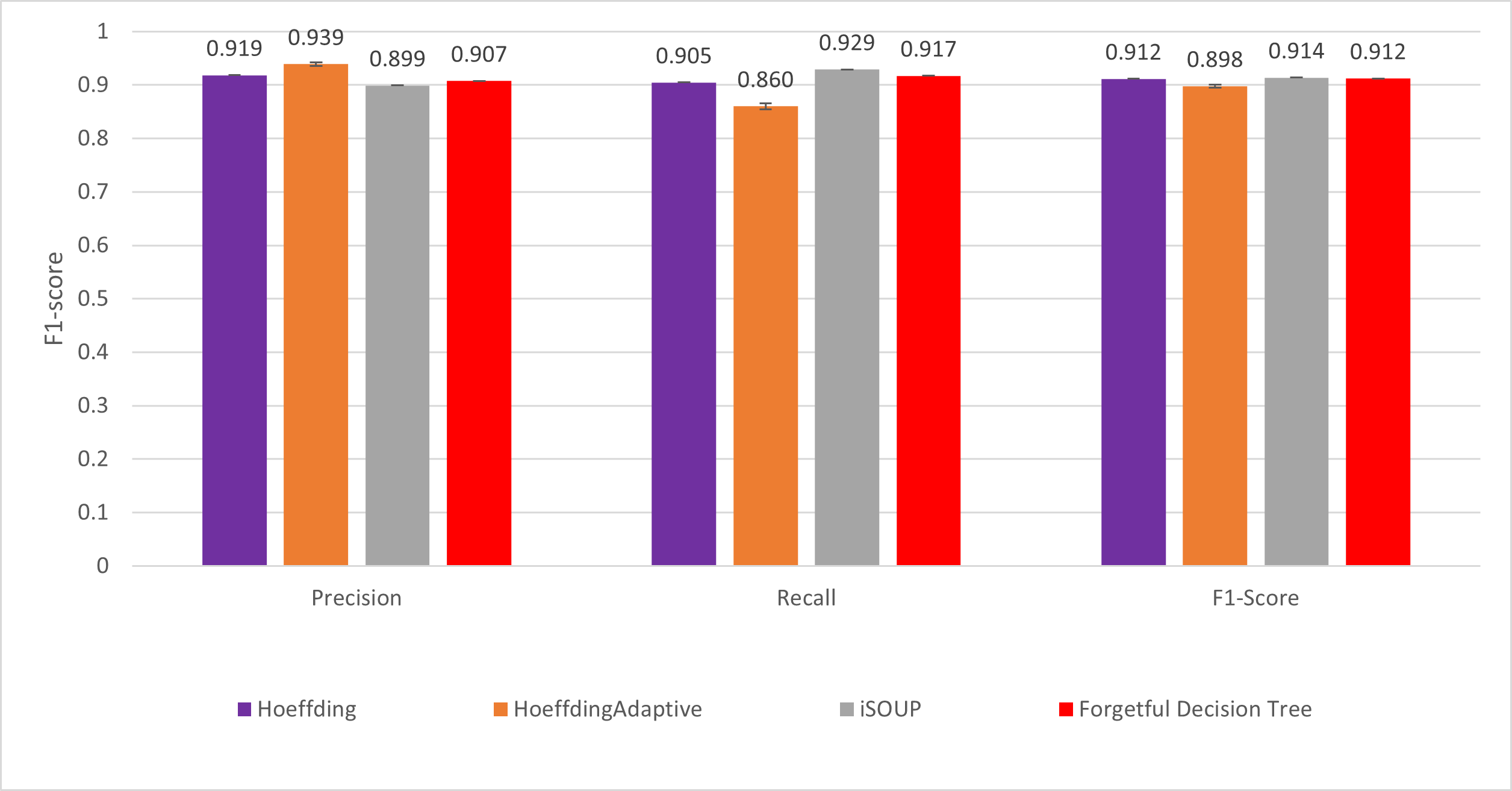}
    \end{figure}

\ \\ Figure \ref{fig:DT_T} compares the time consumption of different incremental decision trees. For all datasets, the Forgetful Decision Tree takes  at least three times less time than the other incremental methods. 

\ \\ Figure \ref{fig:DT_Acc} compares the accuracy of different incremental decision trees. For all datasets, the Forgetful Decision Tree is as accurate or more accurate than other incremental methods.

\ \\ Figure \ref{fig:DT_F1} compares the precision, recall, and F1-score of different incremental decision trees. Because these metrics are not appropriate for other datasets, we  use them only on the phishing dataset. The precision and recalls vary. For example, the Hoeffding Adaptive Tree has a better precision but a worse recall than the Forgetful Decision Tree, while the iSOUP tree has a better recall but a worse precision. Overall,  the Forgetful Decision Tree has a similar F1-score to the other incremental methods.
\ \\
\ \\

\subsection {Quality and Time Performance of Forgetful Random Forest}\label{exp:RF}

    \begin{figure}
        \centering
        \caption{Time Consumption of Random Forests: As can be seen on this logarithmic scale, the Forgetful Random Forest without bagging is at least $24$ times faster than the Adaptive Random Forest. The Forgetful Random Forest with bagging is at least $2.5$ times faster than Adaptive Random Forest.}
        \label{fig:RF_T}
        \includegraphics[width=0.9\textwidth]{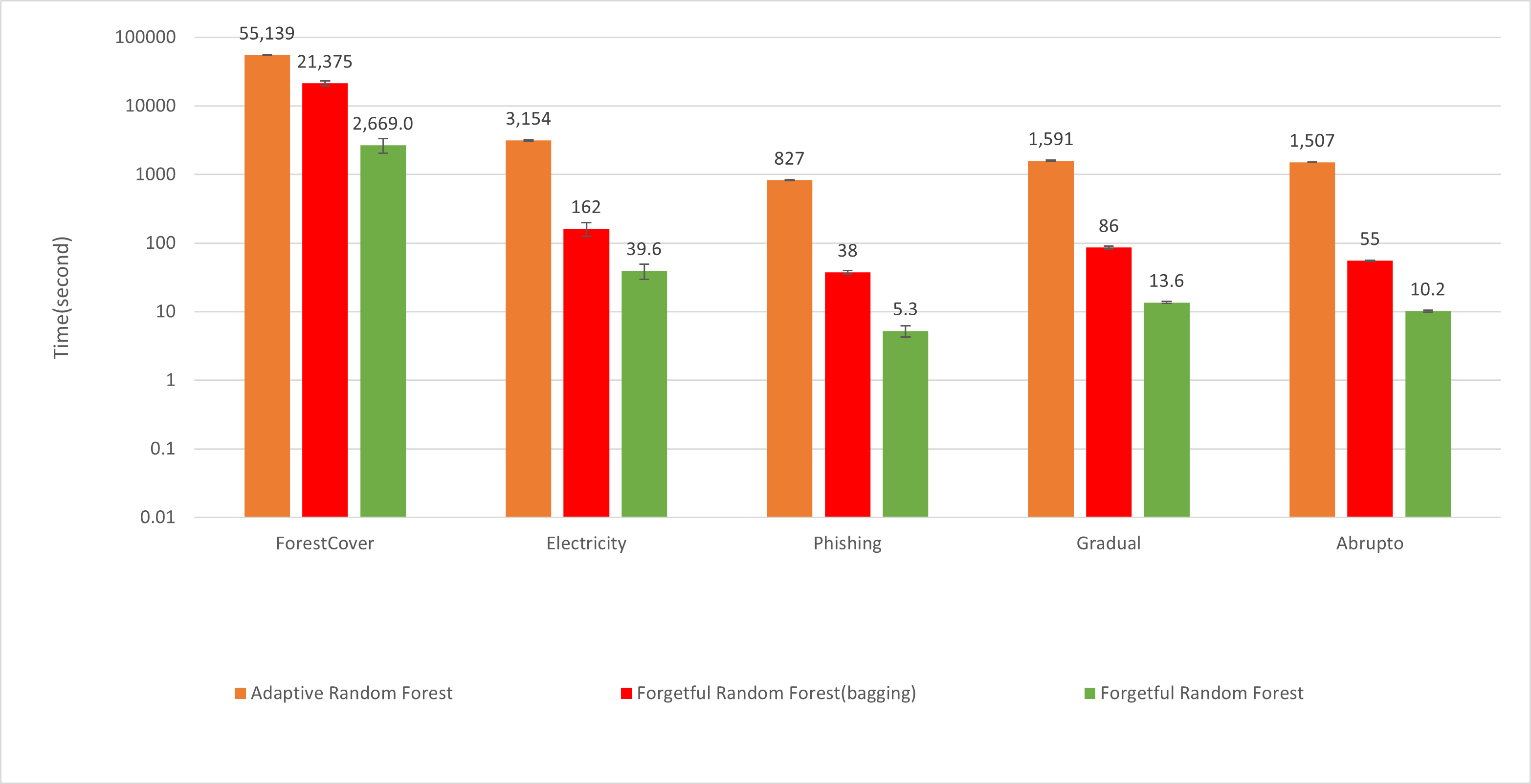}
    \end{figure}
    \begin{figure}
        \centering
        \caption{Accuracy of Random Forests: Without bagging, the Forgetful Random Forest  is slightly less accurate (at most $2\%$)  than the Adaptive Random Forest.  With bagging, the Forgetful Random Forest  has a similar accuracy to the Adaptive Random Forest}
        \label{fig:RF_Acc}
        \includegraphics[width=0.9\textwidth]{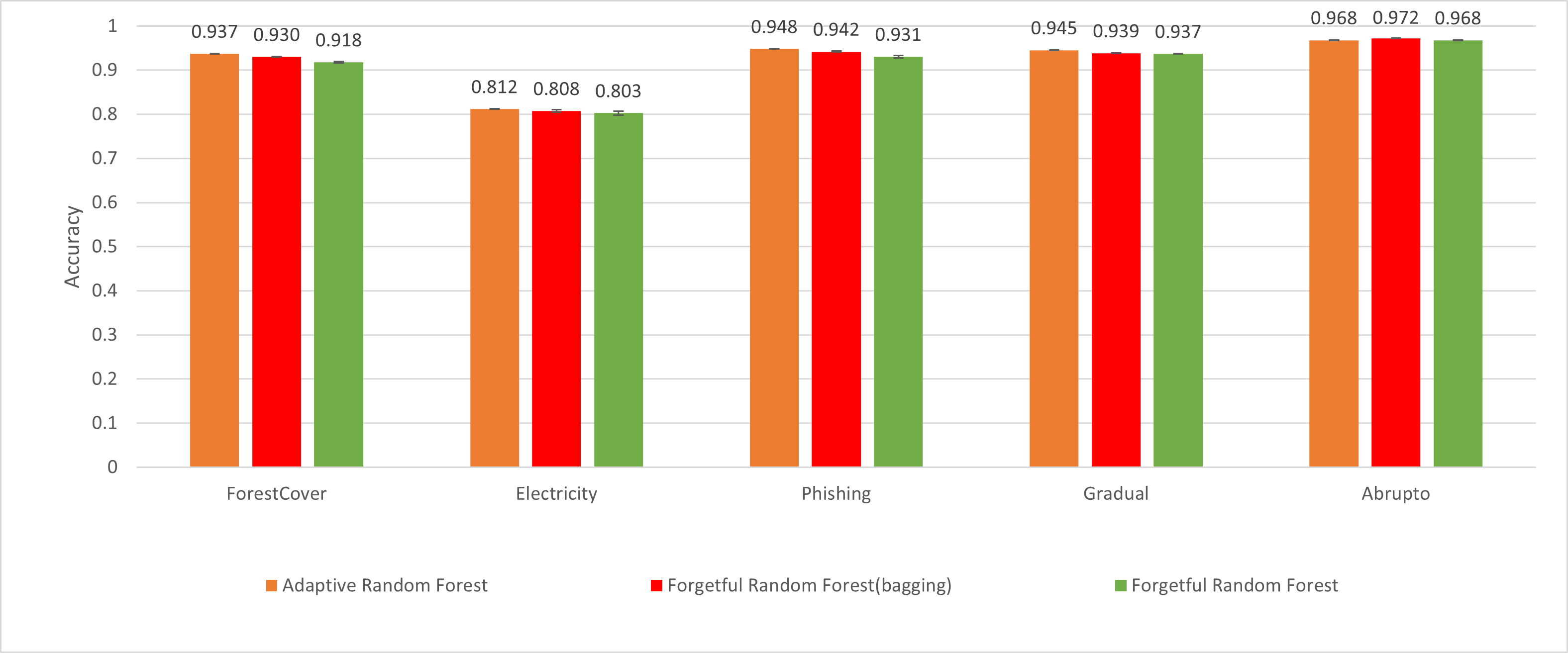}
    \end{figure}
    \begin{figure}
        \centering
        \caption{F1-score of Random Forests: the Forgetful Random Forest without bagging has a lower precision, recall, and F1-score (by $0.02$) compared to the Adaptive Random Forest. The  Forgetful Random Forest with bagging also has a  F1-score (by $0.01$) but a similar precision  to the Adaptive Random Forest.}
        \label{fig:RF_F1}
        \includegraphics[width=0.9\textwidth]{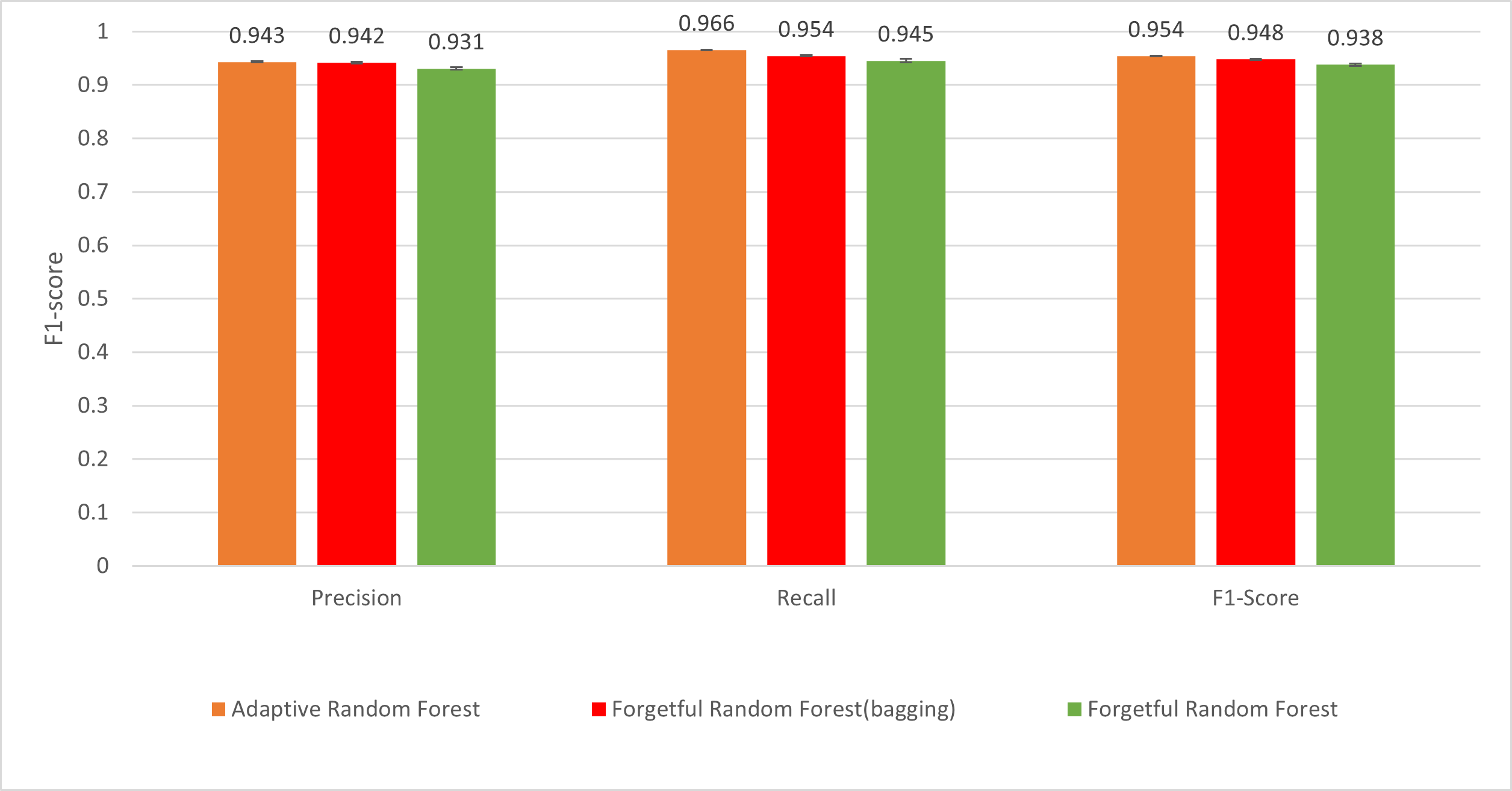}
    \end{figure}


\ \\ Figure \ref{fig:RF_T} compares the time performance of maintaining different random forests. From this figure, we observe that the Forgetful Random Forest without bagging is the fastest algorithm. In particular, it is at least $24$ times faster than the Adaptive Random Forest. The Forgetful Random Forest with bagging is about $10$ times slower than without bagging, but it's still $2.5$ times faster than the Adaptive Random Forest.

\ \\ Figure \ref{fig:RF_Acc}  compares the accuracy of  different random forests. From these figures, we observe that the Forgetful Random Forest without bagging is slightly less accurate than the Adaptive Random Forest (by at most $2\%$). By contrast, the Forgetful Random Forest with bagging has a similar accuracy compared to the Adaptive Random Forest. Sometimes the loss of accuracy might be worth it if the streaming data enters at a high enough rate.

\ \\ Figure \ref{fig:RF_Acc}  compares the precision, recall, and F1-score of training different random forests when these evaluations are appropriate. From these figures, we observe that the Forgetful Random Forest without bagging has a lower precision, recall, and F1-score than the Adaptive Random Forest (by at most $0.02$). By contrast, the Forgetful Random Forest with bagging has a similar precision but a lower recall and F1-score (by at most $0.01$) compared to the Adaptive Random Forest.

\section {Conclusion}

\ \\Forgetful Decision Trees and Forgetful Random Forests constitute simple new fast and accurate incremental data structure algorithms. We have found that

\begin{itemize}
\item 
The Forgetful Decision Tree is at least three times faster and at least as accurate as  state-of-the-art incremental Decision Tree algorithms for a variety of concept drift datasets.  When the precision, recall, and F1-score are appropriate, the Forgetful Decision Tree has  a similar F1-score as state-of-the-art incremental Decision Tree algorithms. 
\item
While Forgetful Random Forest without bagging is at least $24$ times faster than state-of-the-art incremental Random Forest algorithms, it is less accurate by at most $2\%$.
\item
By contrast, the Forgetful Random Forest with bagging has a similar accuracy as the state-of-the-art incremental Random Forest algorithms for concept drift datasets and is $2.5$ times as fast as the Adaptive Random Forest. 
\item
At a conceptual level, our experiments show that it helps to  set parameter values based on changes in accuracy. We do this for $currentParams.rSize$ (retained data), $maxHeight$ (of decision trees), $currentParams.iRate$ (increase rate of $currentParams.rSize$), and the number of features to consider at each decision tree in the Forgetful Random Forests. 

\end{itemize}

\ \\In summary, forgetful data structures  speed up traditional decision trees and random forests and help them adapt to  concept drift. Further we have observed that bagging can increase accuracy but at a substantial loss of performance. The most pressing question for future work is whether some other method can be combined with forgetfulness to increase accuracy at less cost in a streaming concept drift setting.




\bibliography{Incremental.bbl}

\begin{thebibliography}{10}

\bibitem{Cunningham2008}
P{\'a}draig Cunningham, Matthieu Cord, and Sarah~Jane Delany.
\newblock {\em Supervised Learning}, pages 21--49.
\newblock Springer Berlin Heidelberg, Berlin, Heidelberg, 2008.

\bibitem{DeepLearning}
Yann LeCun, Y.~Bengio, and Geoffrey Hinton.
\newblock Deep learning.
\newblock {\em Nature}, 521:436--44, 05 2015.

\bibitem{russel2010}
Stuart Russell and Peter Norvig.
\newblock {\em Artificial Intelligence: A Modern Approach}.
\newblock Prentice Hall, 3 edition, 2010.

\bibitem{IncrementalNN}
Robi Polikar, L.~Upda, S.S. Upda, and Vasant Honavar.
\newblock Learn++: An incremental learning algorithm for supervised neural
  networks.
\newblock {\em Systems, Man, and Cybernetics, Part C: Applications and Reviews,
  IEEE Transactions on}, 31:497 -- 508, 12 2001.

\bibitem{IncrementalSVM}
Chris Diehl and G.~Cauwenberghs.
\newblock Svm incremental learning, adaptation and optimization.
\newblock {\em Proceedings of the International Joint Conference on Neural
  Networks, 2003.}, 4:2685--2690 vol.4, 2003.

\bibitem{CART}
Wei-Yin Loh.
\newblock Classification and regression trees.
\newblock {\em Wiley Interdisciplinary Reviews: Data Mining and Knowledge
  Discovery}, 1:14 -- 23, 01 2011.

\bibitem{VFDT}
Jian Sun, Hongyu Jia, Bo~Hu, Xiao Huang, Hao Zhang, Hai Wan, and Xibin Zhao.
\newblock Speeding up very fast decision tree with low computational cost.
\newblock pages 1272--1278, 7 2020.
\newblock Main track.

\bibitem{Osojnik2017TreebasedMF}
Aljaz Osojnik, P.~Panov, and Saso Dzeroski.
\newblock Tree-based methods for online multi-target regression.
\newblock {\em Journal of Intelligent Information Systems}, 50:315--339, 2017.

\bibitem{10.1145/347090.347107}
Pedro Domingos and Geoff Hulten.
\newblock Mining high-speed data streams.
\newblock In {\em Proceedings of the Sixth ACM SIGKDD International Conference
  on Knowledge Discovery and Data Mining}, KDD '00, page 71–80, New York, NY,
  USA, 2000. Association for Computing Machinery.

\bibitem{doi:10.1080/01621459.1963.10500830}
Wassily Hoeffding.
\newblock Probability inequalities for sums of bounded random variables.
\newblock {\em Journal of the American Statistical Association},
  58(301):13--30, 1963.

\bibitem{10.1007/978-3-642-03915-7_22}
Albert Bifet and Ricard Gavald{\`a}.
\newblock Adaptive learning from evolving data streams.
\newblock In Niall~M. Adams, C{\'e}line Robardet, Arno Siebes, and
  Jean-Fran{\c{c}}ois Boulicaut, editors, {\em Advances in Intelligent Data
  Analysis VIII}, pages 249--260, Berlin, Heidelberg, 2009. Springer Berlin
  Heidelberg.

\bibitem{Ikonomovska2011}
Elena Ikonomovska, Joao Gama, and Saso Dzeroski.
\newblock Learning model trees from evolving data streams.
\newblock volume~23, pages 128--168, 2011.

\bibitem{AdaptiveRF}
Heitor~M. Gomes, Albert Bifet, Jesse Read, Jean~Paul Barddal, Fabr\'{\i}cio
  Enembreck, Bernhard Pfharinger, Geoff Holmes, and Talel Abdessalem.
\newblock Adaptive random forests for evolving data stream classification.
\newblock volume 106, page 1469–1495, USA, oct 2017. Kluwer Academic
  Publishers.

\bibitem{gini}
C.~Gini.
\newblock Concentration and dependency ratios.
\newblock page 769–789, 1997.

\bibitem{10.1007/978-3-642-15880-3_15}
Albert Bifet, Geoff Holmes, and Bernhard Pfahringer.
\newblock Leveraging bagging for evolving data streams.
\newblock In Jos{\'e}~Luis Balc{\'a}zar, Francesco Bonchi, Aristides Gionis,
  and Mich{\`e}le Sebag, editors, {\em Machine Learning and Knowledge Discovery
  in Databases}, pages 135--150, Berlin, Heidelberg, 2010. Springer Berlin
  Heidelberg.

\bibitem{Synthetic}
Jesús López~Lobo.
\newblock {Synthetic datasets for concept drift detection purposes}.
\newblock Harvard Dataverse, 2020.

\bibitem{HT}
Geoff Hulten, Laurie Spencer, and Pedro~M. Domingos.
\newblock Mining time-changing data streams.
\newblock In {\em Knowledge Discovery and Data Mining}, 2001.

\bibitem{JMLR:v12:pedregosa11a}
Fabian Pedregosa, Ga{{\"e}}l Varoquaux, Alexandre Gramfort, Vincent Michel,
  Bertrand Thirion, Olivier Grisel, Mathieu Blondel, Peter Prettenhofer, Ron
  Weiss, Vincent Dubourg, Jake Vanderplas, Alexandre Passos, David Cournapeau,
  Matthieu Brucher, Matthieu Perrot, and {{\'E}}douard Duchesnay.
\newblock Scikit-learn: Machine learning in python.
\newblock {\em Journal of Machine Learning Research}, 12(85):2825--2830, 2011.

\bibitem{covertype}
Dr. Charles W.~Anderson Jock A.~Blackard, Dr. Denis J.~Dean.
\newblock Covertype data set.
\newblock 1998.

\bibitem{Electricity}
A.~Bifet M.~Harries, J.~Gama.
\newblock 2009.

\bibitem{Phishing}
Tegjyot~Singh Sethi and Mehmed Kantardzic.
\newblock On the reliable detection of concept drift from streaming unlabeled
  data.
\newblock arXiv, 2017.

\end{thebibliography}
\bibliographystyle{unsrt}  

\end{document}